\documentclass[]{interact}

\usepackage{graphicx}
\usepackage{lmodern}
\usepackage{amssymb}
\usepackage{graphicx}
\usepackage{amsmath}
\usepackage{amsthm}
\usepackage[caption=false]{subfig}
\usepackage[linesnumbered,ruled,vlined]{algorithm2e}
\DeclareMathOperator*{\argmin}{arg\,min}

\usepackage[utf8]{inputenc}
\theoremstyle{definition}

\usepackage[numbers,sort&compress]{natbib}
\bibpunct[, ]{[}{]}{,}{n}{,}{,}
\renewcommand\bibfont{\fontsize{10}{12}\selectfont}

\begin{document}


\title{Empirical study towards understanding line search approximations for training neural networks}

\author{
\name{Younghwan Chae \textsuperscript{a}\thanks{Younghwan Chae Email: chyohw97@gmail.com} and Daniel N. Wilke\textsuperscript{a}\thanks{Daniel N. Wilke Email: nico.wilke@up.ac.za / wilkedn@gmail.com}}
\affil{\textsuperscript{a}Centre for Asset and Integrity Management (C-AIM), Department of Mechanical
	and Aeronautical Engineering, University of Pretoria, Pretoria, South Africa}
}

\maketitle

\begin{abstract}
Choosing appropriate step sizes is critical for reducing the computational cost of training large-scale neural network models. Mini-batch sub-sampling (MBSS) is often employed for computational tractability. However, MBSS introduces a sampling error, that can manifest as a bias or variance in a line search. This is because MBSS can be performed statically, where the mini-batch is updated only when the search direction changes, or dynamically, where the mini-batch is updated every-time the function is evaluated. Static MBSS results in a smooth loss function along a search direction, reflecting low variance but large bias in the estimated "true" (or full batch) minimum. Conversely, dynamic MBSS results in a point-wise discontinuous function, with computable gradients using backpropagation, along a search direction, reflecting high variance but lower bias in the estimated  "true" (or full batch) minimum. In this study, quadratic line search approximations are considered to study the quality of function and derivative information to construct approximations for dynamic MBSS loss functions. An empirical study is conducted where function and derivative information are enforced in various ways for the quadratic approximations. The results for various neural network problems show that being selective on what information is enforced helps to reduce the variance of predicted step sizes. 
\end{abstract}

\begin{keywords}
Approximation; Response surface; Line search; Neural network; Stochastic Non-Negative Gradient Projection Point; Stochastic gradient; Step size; Learning rate
\end{keywords}

\begin{amscode}
2010 Mathematics Subject Classification codes: 90C15, 90C59, 90C30, 90C26, 90C56
\end{amscode}

\section{Introduction}

Training neural networks on large training sets remain challenging as the effectiveness thereof is determined by step sizes (or learning rates), that are not known upfront. In fact, factors such as the optimizer, data set, scaling and network architecture can significantly influence what is effective. In the midst of these difficulties, we need to search for efficient step sizes with special care \cite{bengio2012practical,Smith2015}, in particular for large-scale problems because choosing an inefficient step size may massively expand the required computational cost, and that then results in energy waste causing unnecessary $ CO_{2} $ emissions \cite{Strubell2019}.

Step size resolution is further complicated by having to resolve them within a stochastic optimization setting \cite{robbins1951stochastic}, since sampling errors are introduced by mini-batch sub-sampling (MBSS). A mini-batch can be updated for every loss function evaluation, often referred to as dynamic MBSS, or only when the search direction is updated referred to as static MBSS. For the latter, the mini-batch therefore remains fixed for loss function evaluations along a search direction. Consequently, information sampled using static mini-batches reflects higher bias and lower variance than dynamic mini-batches, that have lower bias at the cost of higher variance due to the frequent update of the mini-batches  \cite{Kafka2019}. Variance reduction techniques \cite{Wang2013, Johnson2013, Xiao2014, Shang2018} have aimed to reduce sampling errors, but are unable to eliminate them. Albeit, MBSS has enabled the training of numerous machine learning problems, which could not be trained using full-batch sampling \cite{Masters2018}, and remains the industry norm for neural network training. 

It is therefore not surprising that numerous strategies for selecting efficient step sizes within a stochastic setting have been introduced. These include scheduling methods such as cyclical learning rates \cite{Smith2015} and cosine annealing \cite{Loshchilov2016}. However, these often require new hyper-parameters to be selected. Secondly, we have adaptive learning rate methods such as  \textsc{AdaDelta} \cite{Zeiler2012},  \textsc{Adam} \cite{Kingma2014}, which resolve learning rates based on diagonal Hessian approximations. However, these including \textsc{RMSProp} \cite{Tieleman2012}, generalize poorly compared to the stochastic gradient descent (SGD) for some cases \cite{NIPS2017_7003}. Lastly, line searches, which estimate the optimum along a given search direction often using sequential evaluations of the loss function, have proven to be mostly successful using static mini-batches \cite{Bollapragada2017}.

Attempts to resolve learning rates for dynamic MBSS loss functions are limited to
a probabilistic line search \cite{Mahsereci2017a} and an inexact gradient-only line search (GOLS-I) \cite{Kafka2019}. The probabilistic line search used both function value and directional derivative information to construct a Gaussian process surrogate \cite{Mahsereci2017a}, while GOLS-I conducts direct optimization of the loss function along a search direction by locating sign changes in the 
directional derivative  towards finding stochastic non-negative gradient projection points (NN-GPP). Although GOLS-I uses less information it achieves similar or better performance when compared to the probabilistic line search  \cite{Kafka2019}.

This naturally raises the question of the usefulness of function value and directional derivative information when constructing approximations, which is subsequently the focus of this study. We aim to investigate the quality of function value and directional derivative information in the construction of quadratic line search approximations. Specifically, we construct a 1D quadratic approximation in five ways using
\begin{enumerate}
	\item only function values (one approximation)
	\item both function values and directional derivatives (three approximations)
	\item only directional derivative information (one approximation)
\end{enumerate}
We will investigate the quality of the different pieces of information using a quadratic approximation, which is the simplest polynomial function with one extremum. This implies a linear approximation for the gradient-only surrogate or approximation. Although quadratic approximations may seem simplistic, they are popular in practice, which include second order and state-of-art adaptive learning rate methods such as \textsc{AdaDelta} and \textsc{Adam}. These effectively solve highly non-linear and ill-conditioned problems when compared to first order approaches \cite{Bordes2009, Bottou2018}.

As a simple quadratic approximation may introduce bias into the step size predictions, we focus on the variance of the predicted step sizes to assess the usefulness of the approximations. In particular, we consider logistic regression and deeper networks using Wisconsin diagnostic breast cancer (WDBC) \cite{Wolberg2011}, MNIST \cite{LeCun1998} and CIFAR-10 \cite{Krizhevsky2009a} datasets in our investigations to assess the usefulness of function values and directional derivatives to construct approximations. Our primary finding is that directional derivative information at the initial point is essential to reduce the variance in predicted step sizes. By omitting additional information, in particular function values, variance in predicted step sizes may be further reduced, which is congruent with our observation of the probabilistic line search and GOLS-I.

\section{Related work}
\subsection{Mini-batch sub-sampling}
The neural network loss function $\mathcal{L}(\boldsymbol{x})$, of $n$ weights $\boldsymbol{x}\in\mathcal{R}^n$, computed using the full batch $ \mathcal{B}_{f} $ of  $ M $ samples, is defined as the average loss of the individual sample losses $ \ell $ over all samples (\ref{loss}). $ \mathcal{L}(\boldsymbol{x})$ can be approximated using mini-batch sampling \cite{robbins1951stochastic} by randomly selecting mini-batches $ \mathcal{B} \in \mathcal{B}_{f} $ of $ m $ samples, where $ 1\leq m \leq M $, usually $ m<<M $. The full batch loss $ \mathcal{L} $ and mini-batch loss $\hat{\mathcal{L}}$ are given by
\begin{equation}\label{loss}
\mathcal{L}(\boldsymbol{x}):= \dfrac{1}{M} \sum_{i = 1}^{M}\ell(\boldsymbol{x})_{\{s\}_{i}}\approx \hat{\mathcal{L}} := \dfrac{1}{m}\sum_{j = 1}^{m} \ell(\boldsymbol{x})_{\{s\}_{j}},
\end{equation}
where $ {\{s\}}_{i} \in \mathcal{B}_{f}$ and $ {\{s\}}_{j} \in \mathcal{B}$, denote the individual samples from the full sample set $ \mathcal{B}_{f} $ and mini-batch set $ \mathcal{B} $, respectively. Equivalently, the full batch gradient $ \bm{\nabla\mathcal{L}} $ and mini-batch gradient $ \bm{\nabla}\hat{\bm{\mathcal{L}}} $ are given by
\begin{equation}\label{mini-grad}
\bm{\nabla\mathcal{L}}(\boldsymbol{x}):= \dfrac{1}{M} \sum_{i = 1}^{M}\bm{\nabla\ell}(\boldsymbol{x})_{\{s\}_{i}}\approx \bm{\nabla}\hat{\bm{\mathcal{L}}} := \dfrac{1}{m}\sum_{j = 1}^{m} \bm{\nabla\ell}(\boldsymbol{x})_{\{s\}_{j}}
\end{equation}
where the gradient of an individual sample loss $ \ell $ is denoted by $ \bm{\nabla\ell} $.  There are popular techniques related to MBSS; importance sampling, which prioritizes samples, that contribute more information \cite{Katharopoulos2018,JMLR:v19:16-241,NIPS2018_7957} and adaptive sampling, which increases the mini-batch size using an inner product test \cite{Bollapragada2017,Friedlander2011} to reduce gradient variance from mini-batch sampling thereby  accelerating the training process. Although improvements have been made using more advanced sampling strategies, the variance due to sampling error can never be eliminated completely. However, random sub-sampling remains the industry standard, which is also the approach we consider in this study to avoid distracting from our investigation.

\subsection{Line searches: static vs. dynamic mini-batch sub-sampling}
\begin{figure}
	\centering
	\subfloat[Sampling structure]{%
		\resizebox*{7cm}{!}{\includegraphics{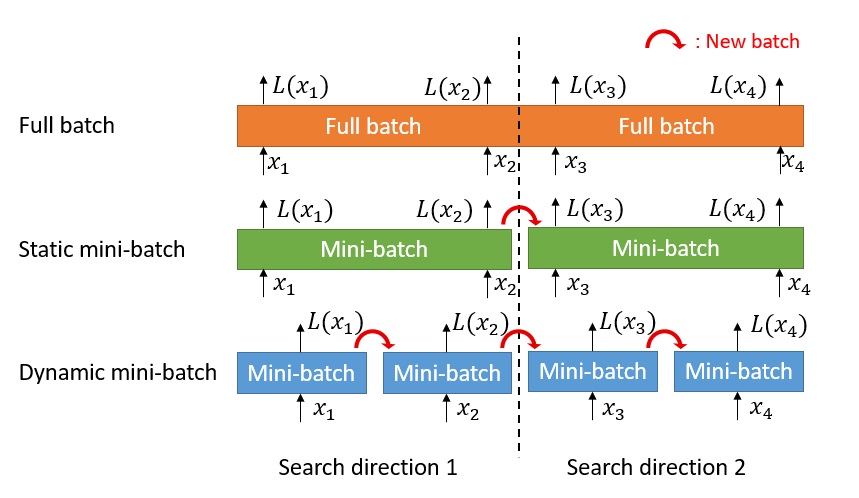}}}\hspace{5pt}
	\subfloat[Distribution of computed losses]{%
		\resizebox*{7cm}{!}{\includegraphics{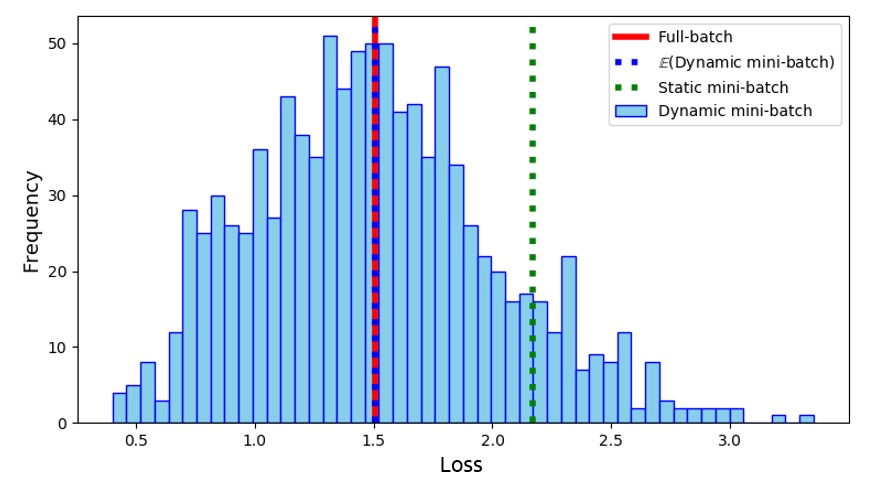}}}
	\caption{(a) An illustration of the sampling structure within an algorithm for full-batch, static mini-batch and dynamic MBSS. Static MBSS keeps a chosen mini-batch fixed for every function evaluation along a search direction. Conversely, Dynamic MBSS re-samples the mini-batch for every function evaluation. (b) An example of the distribution of $ \mathcal{L} $ and $ \hat{\mathcal{L}} $ using different sampling methods.} \label{staticdynamic}
\end{figure}

Static MBSS refers to sub-sampling where the mini-batch only updates every-time a new search direction is computed, whereas dynamic MBSS indicated to sub-sampling where the mini-batch is updated every-time the loss function is computed, as shown in Figure~\ref{staticdynamic}(a). The distributions of the losses computed using full-batch sampling, dynamic MBSS and static MBSS at a chosen point are shown in Figure~\ref{staticdynamic}(b). It shows that the expected value of the losses from dynamic MBSS is close to the full batch solution. This is akin to the bias-variance trade-off, where static MBSS has large bias and small variance, whereas dynamic MBSS has small bias and large variance. Considering sub-gradient methods or stochastic gradient descent, there is no distinction between static and dynamic sub-sampling as every-time the loss function is computed the search direction is updated. However, within the context of line searches it is essential to distinguish between static and dynamic sub-sampled loss functions, since line searches conduct multiple loss function evaluations for a given search direction \cite{nocedal2006numerical, Snyman2018}. 

A typical line search aims to estimate the minimizer \cite{nocedal2006numerical}
\begin{equation}\label{ls}
\alpha^* = \argmin_\alpha\hat{\mathcal{L}} \boldsymbol(\boldsymbol{x}+\alpha\boldsymbol{d}),
\end{equation}
along a descent direction $\boldsymbol{d} \in \mathcal{R}^n$ from the current point $\boldsymbol{x} \in \mathcal{R}^n$, where $\hat{\mathcal{L}}$ is computed using uniformly random mini-batches.

\begin{figure}
	\centering
	\subfloat[Discontinuity of loss and directional derivative]{%
		\resizebox*{7cm}{!}{\includegraphics{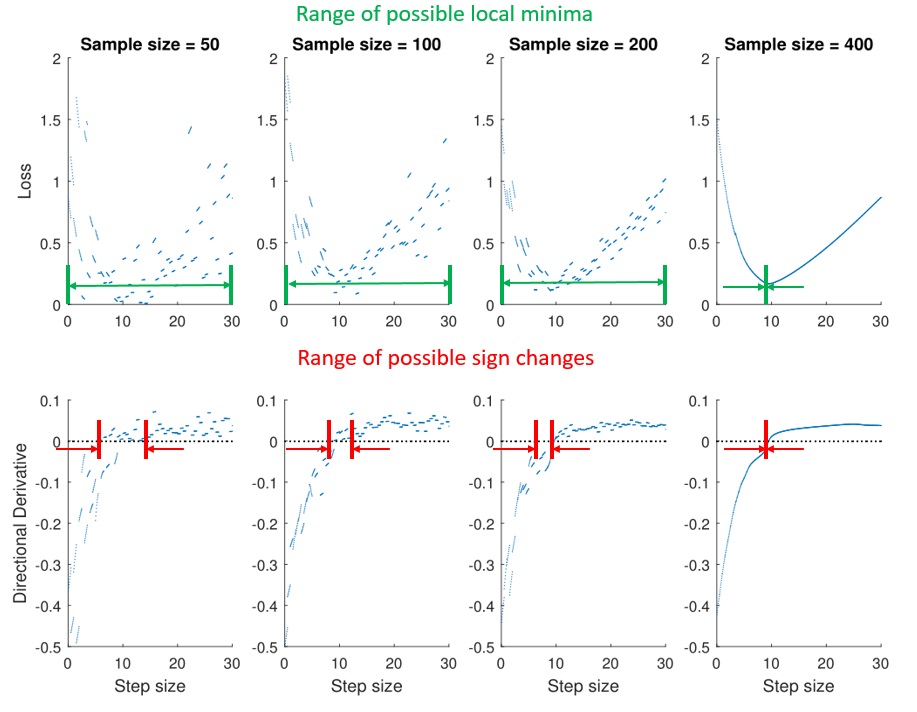}}}\hspace{5pt}
	\subfloat[Distribution of solutions]{%
		\resizebox*{6cm}{!}{\includegraphics{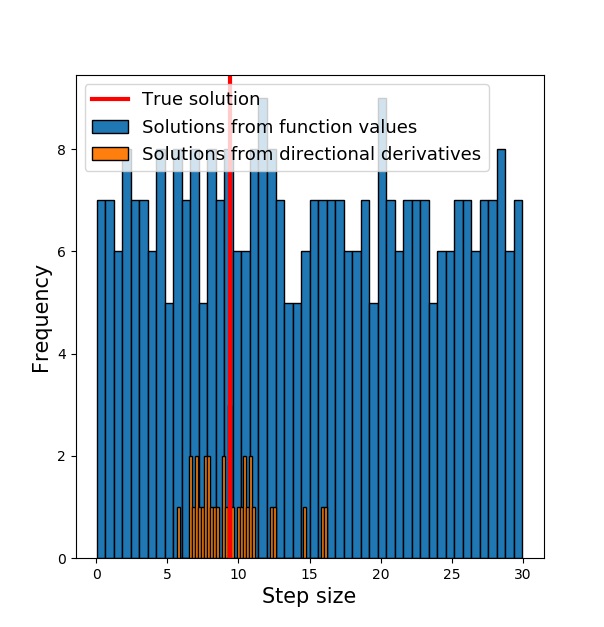}}}
	\caption{(a) Plots of discontinuous loss (top) and directional derivative (bottom) functions  from a logistic regression problem using the breast cancer dataset with various mini-batch sizes (50, 100, 200 and the full training batch of 400). Each range of possible local minimum locations when observing the loss values is indicated in green and the range of possible sign change locations when observing only directional derivative information is indicated in red. (b) The respective distributions of solutions computed using either function values or directional derivatives.} \label{discontin}
\end{figure}

However, the nature of $\hat{\mathcal{L}}$ is significantly affected by whether we use static or dynamic MBSS. As pointed out, for static MBSS, the minimizer $\alpha^*$ is resolved using a fixed mini-batch, i.e. the sampling error is constant and the resulting loss function $\hat{\mathcal{L}}$'s smoothness and continuity is dictated only by the neural network architecture and selection of activation functions. In contrast, when $\hat{\mathcal{L}}$ is computed using dynamic MBSS, the sampling error changes with every evaluation resulting in a guaranteed point-wise discontinuous loss function 
$\hat{\mathcal{L}}$, irrespective of the neural network architecture or activation functions. This is shown in Figure~\ref{discontin}(a) for logistic regression on the breast cancer dataset for three mini-batch sizes, 50, 100 and 200, and the full batch of 400.

This implies that for the static MBSS, the step size $ \alpha $ is chosen based on a single mini-batch $ \mathcal{B} $, which may introduce large bias in the step size estimate. However, the sampling error remains constant for every evaluation along the search direction essentially eliminating any variance due to sampling errors when evaluating the loss function. In turn, for dynamic MBSS, the sampling error changes for every function evaluation, which may introduce high variance in function and gradient estimates due to changes in sampling error for every evaluation. The benefit being that a higher throughput of data is achieved, which in turn reduces the bias of the step size estimate.

If we were to consider searching for minimizers for the dynamic MBSS loss functions $ \hat{\mathcal{L}} $ in Figure~\ref{discontin}(a), it is clear that full batches are required. The step variance for potential minimizers, as indicated in green on the top rows, highlights this complication. In fact, the range of local minima does not significantly decrease with batch size, unless the full batch is considered. In turn, when we consider finding sign changes from negative to positive in the directional derivative, following a gradient-only definition for optimality as outlined by \cite{Snyman2018, Kafka2019}, the step size variance is significantly reduced as indicated in red on the bottom row. It is also clear that the domain decreases with an increase in batch size. For the full batch, the information is equivalent when isolating local minima. Figure~\ref{discontin}(b) shows the respective distributions of solutions computed using function values and directional derivatives. It is evident that even when considering dynamic MBSS loss functions the bias and variance between the function values and directional derivatives vary significantly, that may in turn significantly influence the quality of approximations constructed to estimate optimal step sizes. 	

\subsection{Line search algorithms}
Adaptive learning rate methods \cite{Gulcehre2014,Gulcehre2017, Schaul2012,Schaul2013} tend to resolve both the search direction and step size due to component-wise updating. Conversely, line searches aim to estimate the minimizer $\alpha^*$ along a given search direction to resolve the step size, as outlined in \eqref{ls}. However, line searches have gained limited momentum for static MBSS loss functions \cite{Bollapragada2017,Friedlander2011}, while resolving step sizes for dynamic MBSS loss functions are still in its infancy and limited to three investigations.

Firstly, based on Bayesian optimization, a Gaussian process surrogate model was proposed by \cite{Mahsereci2017a} to estimate the expected local minima by satisfying the Wolfe condition \cite{wolfe1969convergence, wolfe1971convergence} in a probabilistic sense for stochastic gradient descent. In turn, \cite{wills2017construction} proposed a probabilistic quasi-Newton method. Recently, \cite{Kafka2019} developed an inexact gradient-only line search (GOLS-I), that explicitly evaluates directional derivatives to locate sign changes from negative to positive. 

GOLS-I is based on the principles underlying the gradient-only optimization problem, that aim to resolve a Non-Negative Associated Gradient Projection Point (NN-GPP) $\boldsymbol{x}_{nngpp}$ \cite{wilke2013gradient, Snyman2018}, which is given by 
\begin{equation}\label{key}
\boldsymbol{d}^{T}\cdot \bm{\nabla} \hat{\bm{\mathcal{L}}}(\boldsymbol{x}_{nngpp} + \alpha\boldsymbol{d}) \geq 0, \quad \forall \|\boldsymbol{d} \in \mathbb{R}^{D}\|_{2} = 1, \quad \forall\alpha \in (0, \alpha_{max}].
\end{equation}
For dynamic MBSS loss functions, instead of locating $\boldsymbol{x}_{nngpp}$, the definition for NN-GPP was recently extended by \cite{Kafka2019} to Stochastic NN-GPP (SNN-GPP), that defines candidate optimal points in a ball due to the stochastic nature and high variance of the sampled information. This 1-D ball is evident in the plots below in Figure~\ref{discontin}(a), which shrinks with increasing batch size. 

\subsection{Approximation functions}
Approximation (or surrogate) functions are often used in engineering to analyze, and optimize expensive black-box problems  \cite{Vu2017, Bhosekar2018}. These approximations are usually constructed by regression to estimate function values  at sampled locations, as gradients are often not available for engineering problems \cite{Regis2013}. In cases, where gradients are available, it has been shown to improve the quality of surrogate models through gradient-enhanced Kriging \cite{Chen2019a,Laurent2019a}. In neural network training, analytical gradient information is explicitly computable through back-propagation \cite{Rumelhart1986}.

Approximation functions are susceptible to the curse of dimensionality \cite{Kubicek2015}, 
making them unsuited for approximations of high dimensional problems like neural network training. Albeit, \cite{Mahsereci2017a} demonstrated that probabilistic surrogates are well-suited to approximate 1-D line searches. In addition, \cite{Wilke2016, Snyman2018} showed that smooth approximations can be constructed of discontinuous functions using only gradient information. This enables us to consider the construction of 1D quadratic approximations by enforcing different information such as function value and directional derivatives, at different locations. In particular, we investigate the variance of step sizes resolved using five 1D quadratic approximations in Section~\ref{sec:modelss}.



\section{Enforcing selective information to construct 1D quadratic approximations}\label{sec:modelss}
In the construction of 1D quadratic approximations, selected information is enforced at different points along the descent direction. Specifically, we consider the following five approximations using function value and/or directional derivative information at $\alpha_0 = 0$, $\alpha_1>0$ and $\alpha_2>0$, with $\alpha_1 > \alpha_2$: 
\begin{enumerate}
	\item function-value-only (f-f-f): the function values at $ \alpha_{0} $, $ \alpha_{1} $ and $ \alpha_{2} $, respectively denoted by $ f_{0} $, $ f_{1} $ and $ f_{2} $;
	\item mixed (fg-f): the function value and directional derivative at $ \alpha_{0} $ and the function value at $ \alpha_{1}$, respectively denoted by $ f_{0} $, $ f'_{0} $ and $ f_{1} $; 
	\item mixed (f-fg): the function value at $ \alpha_{0} $ and the function value and directional derivative  at $ \alpha_{1}$, respectively denoted by $ f_{0} $, $ f_{1} $ and $ f'_{1} $;
	\item mixed (fg-fg): the function value and directional derivative at $ \alpha_{0} $ and the function value at $ \alpha_{1}$, respectively denoted by $ f_{0} $, $ f'_{0} $, $ f_{1} $ and $ f'_{1} $;
	\item derivative-only (g-g): the directional derivative measured at both $ \alpha_{0} $ and $ \alpha_{1}$, respectively denoted by $ f'_{0} $ and $ f'_{1} $. 
\end{enumerate}

\begin{figure}
	\centering
	\includegraphics[width=0.7\linewidth]{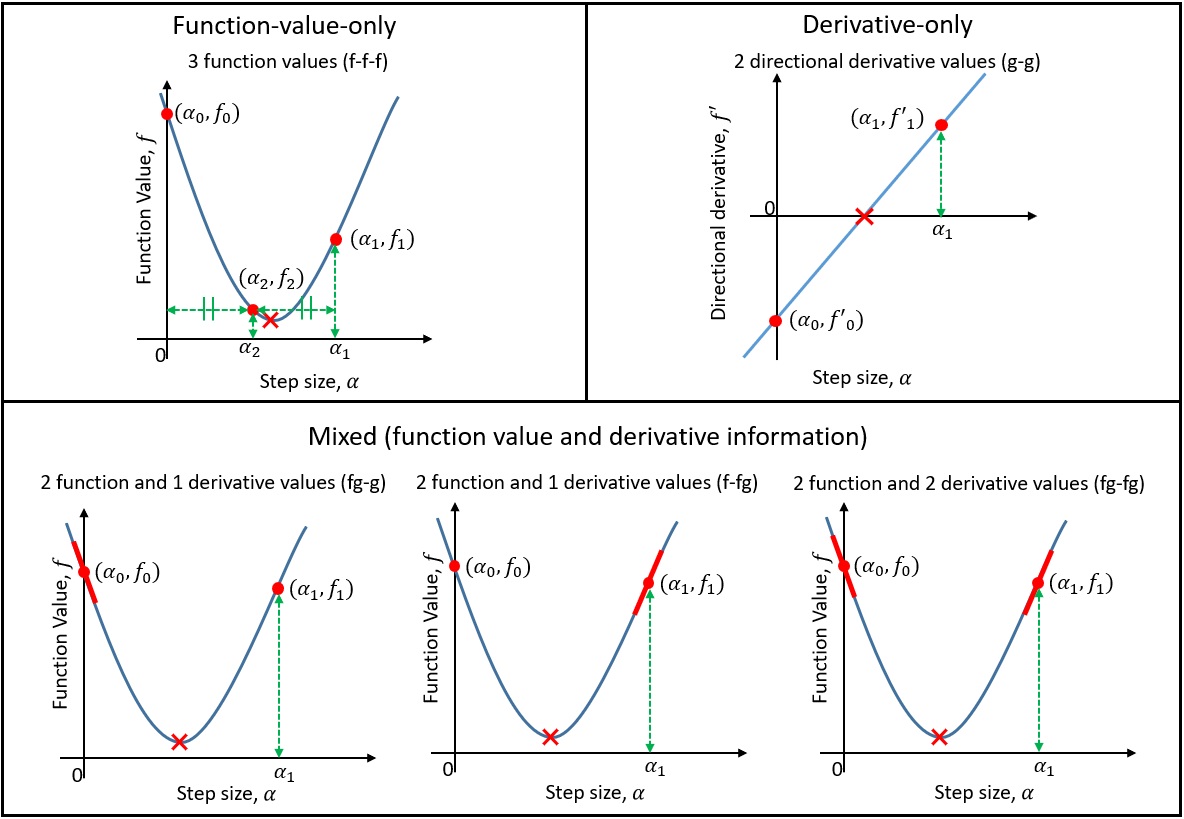}
	\caption{Illustration of quadratic approximations using different types of information; (top left) the function-value-only approximation (f-f-f), (bottom right) the mixed approximation with the directional derivative at $ \alpha_{0} $ (fg-f), (bottom middle) the mixed approximation with the directional derivative at $ \alpha_{1} $ (f-fg), (bottom right) the mixed approximation with the directional derivatives at both $ \alpha_{0} $ and $ \alpha_{1} $ (fg-fg) and (top right) derivative-only approximation (g-g).}
	\label{threetypes}
\end{figure}

Given a multivariate dynamic MBSS  loss function $ \hat{\mathcal{L}}:(\boldsymbol{x}\in \mathbb{R}^{n}) \rightarrow \mathbb{R} $ to be approximated $ \hat{\mathcal{L}} $ along the search direction $ \boldsymbol{d}\in \mathbb{R}^{n} $ by a 1-D quadratic function $ \tilde{f}(\alpha) $ 
\begin{equation}\label{quad_func}
\hat{\mathcal{L}}(\boldsymbol{x} + \alpha\boldsymbol{d}) = f(\alpha) \approx \tilde{f}(\alpha) = k_{1}\alpha^{2} + k_{2}\alpha + k_{3}.
\end{equation}
The directional derivative $ \bm{\nabla}\hat{\bm{\mathcal{L}}}^\textrm{T}(\boldsymbol{x} + \alpha\boldsymbol{d})\boldsymbol{d}$ is approximated by the following linear function
\begin{equation}\label{lin_func}
\bm{\nabla}\hat{\bm{\mathcal{L}}}^\textrm{T}(\boldsymbol{x} + \alpha\boldsymbol{d})\boldsymbol{d} = f'(\alpha) \approx \tilde{f}'(\alpha) =  2k_{1}\alpha+k_{2}.
\end{equation}
It is evident that the approximation $ \tilde{f} $ required at least three pieces of independent information to resolve $k_1, k_2$ and $k_3$, while $ f'(\alpha) $ requires only two pieces of independent information to resolve $k_1$ and $k_2$. The five approximations with the required  information (red circle indicates respective function value and red slope the derivative of the respective function) to construct each are shown in Figure~\ref{threetypes}. All approximations are well-specified linear systems except for the over-specified mixed (fg-fg) approximation, that  requires a least squares solution to approximate four pieces of information with only three variables.   

Each approximation can be constructed by solving a small linear system, $\boldsymbol{A}_i \boldsymbol{k}_i = \boldsymbol{b}_i,\;i=1,2,3,4,5$, that is obtained by evaluating
\eqref{quad_func} and/or \eqref{lin_func} at selected $\alpha$ values to 
enforce / approximate the respective function or derivative information, with $\boldsymbol{A}_i$
and $\boldsymbol{b}_i$ respectively given by

\begin{equation}\label{fff}
\boldsymbol{A}_{1} =   \begin{bmatrix}
0 &0&1 \\
\alpha_{1}^{2}&\alpha_{1} & 1\\
\alpha_{2}^{2}&\alpha_{2} & 1\\
\end{bmatrix}; \boldsymbol{b}_{1} =  \begin{bmatrix}
\hat{f}_{0} & \hat{f}_{1} & \hat{f}_{2}
\end{bmatrix}^{T}  \\
\end{equation}

\begin{equation} \label{fgf}
\boldsymbol{A}_{2} =   \begin{bmatrix}
0 &0&1 \\
0&1&0\\
\alpha_{1}^{2}&\alpha_{1} & 1\\
\end{bmatrix}; \boldsymbol{b}_{2} =  \begin{bmatrix}
\hat{f}_{0} & \hat{f}'_{0} & \hat{f}_{1}
\end{bmatrix}^{T}
\end{equation}

\begin{equation}\label{ffg}
\boldsymbol{A}_{3} =  \begin{bmatrix}
0 &0&1 \\
\alpha_{1}^{2}&\alpha_{1} & 1\\
2\alpha_{1} & 1&0\\
\end{bmatrix};  \boldsymbol{b}_{3} =  \begin{bmatrix}
\hat{f}_{0} & \hat{f}_{1} & \hat{f}'_{1}
\end{bmatrix}^{T}
\end{equation}

\begin{equation}\label{fgfg}
\boldsymbol{A}_{4} =   \begin{bmatrix}
0 &0&1 \\
0&1&0\\
\alpha_{1}^{2}&\alpha_{1} & 1\\
2\alpha_{1} & 1&0\\
\end{bmatrix}; \boldsymbol{b}_{4} =  \begin{bmatrix}
\hat{f}_{0} & \hat{f}'_{0}& \hat{f}_{1} & \hat{f}'_{1}
\end{bmatrix}^{T}
\end{equation}

\begin{equation}\label{gg}
\boldsymbol{A}_{5} =   \begin{bmatrix}
0 &1 \\
2\alpha_{1} & 1\\
\end{bmatrix} 
; \boldsymbol{b}_{5} =  \begin{bmatrix}
\hat{f}'_{0} & \hat{f}'_{1}
\end{bmatrix}^{T}
\end{equation}

All approximations are interpolation models except for the mixed approximation (fg-fg), which is a regression model. We briefly discuss a number of consequences of using basic approximations, that include approximation bias and extrapolation.

\subsection{Approximation bias}

\begin{figure}
	\centering
	\subfloat[Overshooting]{%
		\resizebox*{5cm}{!}{\includegraphics{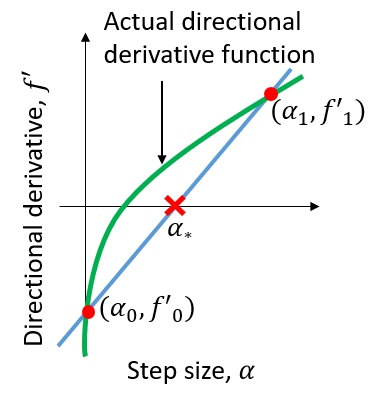}}}\hspace{5pt}
	\subfloat[Undershooting]{%
		\resizebox*{5cm}{!}{\includegraphics{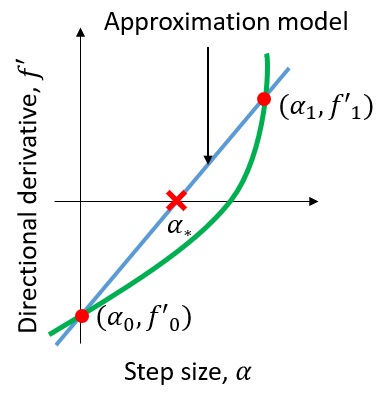}}}
	\caption{Illustration of approximation bias that either (a) overshoot or (b) undershoot the actual solution given by the stationary point for the actual direction derivative in green, due to the mismatch between actual directional derivative and the basic approximation. The red cross indicates the approximated location of the sign change from the derivative-only approximation.} \label{shooting}
\end{figure}

The 1D derivative function for our 1D quadratic approximations is linear, which is the simplest 1D derivative approximation, that allows for variation. The minimum of the 1D quadratic approximation is given by a sign change from negative to positive of the first derivative, or alternatively stated, a zero derivative with positive curvature or second derivative. Given that our approximation of the derivative is basic, and that we are trying to approximate an actual function, that is significantly more complex, our approximations is subject to bias. The potential bias of our approximations are illustrated in Figure~\ref{shooting}, which depicts the directional derivative of the actual function and derivative of our 1D approximation. The solutions (red crosses) given by the zero derivative obtained using the approximation (in blue) may result in either overshooting or undershooting of the actual zero derivative of the actual directional derivative. This depends on the curvature of the actual derivative function (in green). This implies that one can predict the shape of the actual function by observing the resolved step sizes $ \alpha_{*} $. 

It is clear, that approximation bias may result in over or undershooting of the actual solution. In order to assess the quality of the various surrogate approximations, we focus on the variance of the estimates. For this reason, we need to pay careful attention on whether our estimates are interpolation or extrapolated from the approximation, as extrapolation may amplify variance in the estimates.  

\subsection{Extrapolation}
For a 1D quadratic approximation (\ref{quad_func}) or linear derivative approximation (\ref{lin_func}), the curvature $ 2 k_{1} $ is convex if $ k_{1} > 0$ or concave if $ k_{1} < 0 $. In addition, $ \alpha_{0} $ and $ \alpha_{1} $ could both be associated with negative directional derivatives which imply, that the optimum is beyond $\alpha_1$, requiring extrapolation. 
Figure~\ref{cases} depicts three potential cases for solving step sizes using a 1D quadratic approximation: 1)~unbounded extrapolation, 2)~bounded extrapolation and 3)~interpolation. The cases depend on what were evaluated at $ \alpha_{0} $ and $ \alpha_{1} $ as shown in Figure~\ref{cases}(a) and (b) for the function approximation and derivative approximation, respectively. The three possible extrapolation or interpolation events for the approximation-assisted line search is as follows:

\begin{figure}
	\centering
	\subfloat[Function value domain]{%
		\resizebox*{5cm}{!}{\includegraphics{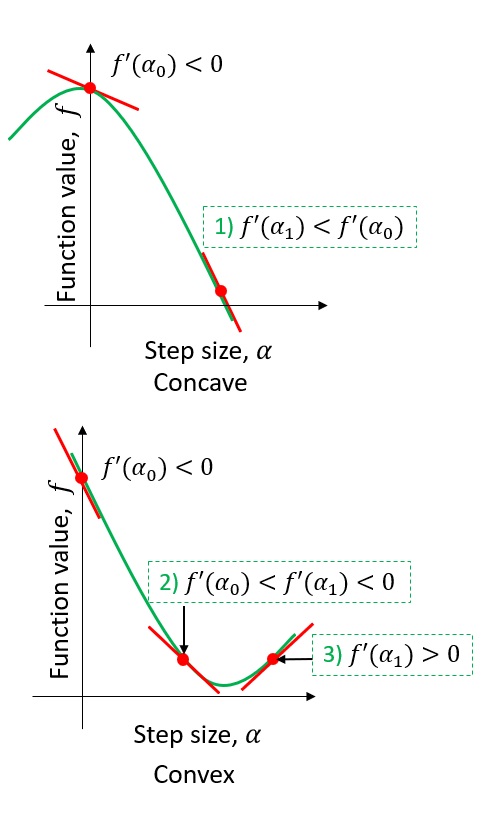}}}\hspace{5pt}
	\subfloat[Directional derivative domain]{%
		\resizebox*{5cm}{!}{\includegraphics{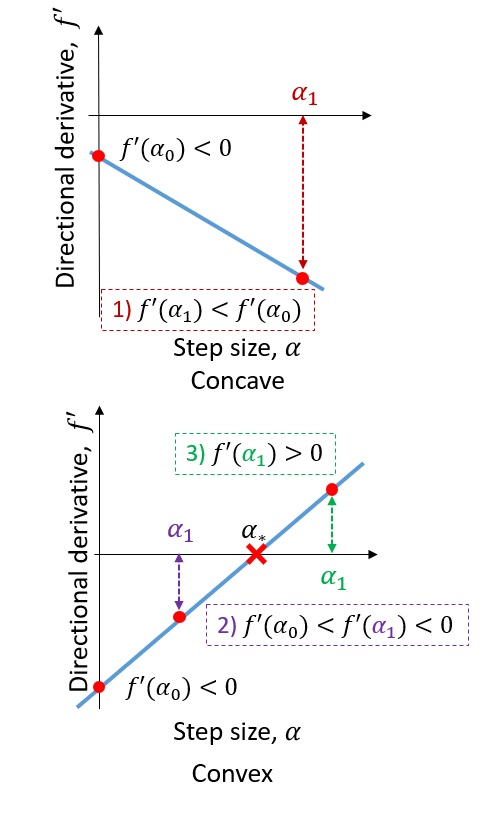}}}
	\caption{Illustration of 1) unbounded extrapolation and 2) bounded extrapolation 3) interpolation  situations for approximations  in (a) function value or (b) directional derivative domain.} \label{cases}
\end{figure}


\begin{enumerate}
	\item If $ f'_{1} < f'_{0} < 0 $, the approximation becomes concave for unbounded extrapolation;
	\item If $ f'_{0} <f'_{1}<0 $, the approximation becomes convex for bounded extrapolation, with local minimum $ \alpha_{*} $ at $ \alpha_{*}> \alpha_{1} $;
	\item If $ f'_{1}> 0  $, the approximation interpolates, and there is a local minimum  $ \alpha_{*} $ at $ \alpha_{0} < \alpha_{*}<\alpha_{1}$.
\end{enumerate}

\begin{figure}
	\centering
	\subfloat[Interpolation]{%
		\resizebox*{5cm}{!}{\includegraphics{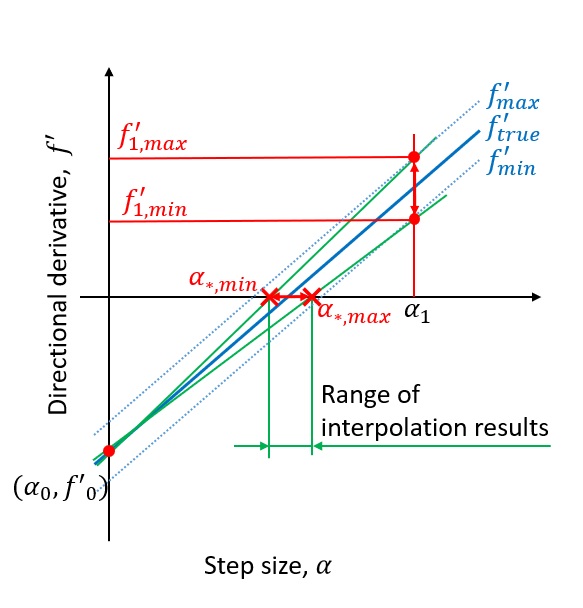}}}\hspace{5pt}
	\subfloat[Bounded extrapolation]{%
		\resizebox*{5cm}{!}{\includegraphics{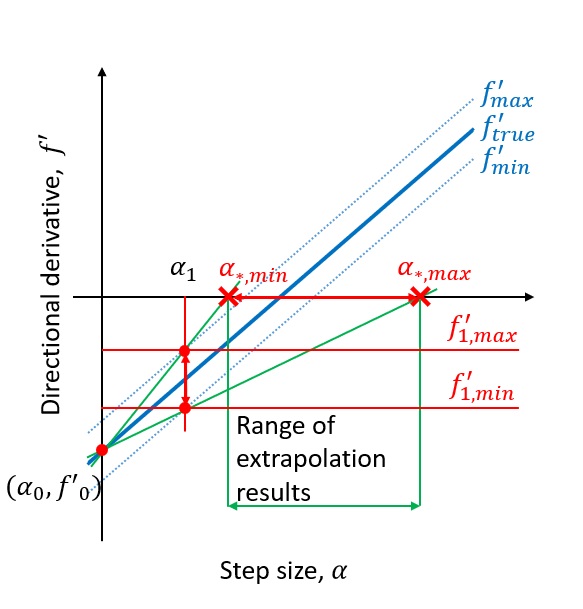}}}
	\caption{Illustration of possible ranges of (a) interpolation and (b) bounded extrapolation results $ \alpha_{*} \in [\alpha_{*,min}, \alpha_{*,max}]$ due to variances in the directional derivatives caused by dynamic MBSS. Assumed is a constant derivative variance independent of initial guess $ \alpha_{1} $, $ f'_{1} \in [f'_{1,min},f'_{1,max}]$.} \label{constvar}
\end{figure}

Some of the consequences of bounded extrapolation estimates compared to interpolation estimates is captured by Figure~\ref{constvar}. We illustrate the ranges of possible solution estimates of $ \alpha_{*} \in [\alpha_{*,min},\alpha_{*,max}] $ computed using (a) interpolation and (b) bounded extrapolation. We assume a constant bounded difference for $ f'_{1}\in [f'_{1,min},f'_{1,max}] $ along the step size $ \alpha $, where the variance in $ f'_{1} $ is introduced by dynamic MBSS. Each of the green lines illustrates the approximations constructed using either $(f'_{0}, f'_{1,min}) $ or $(f'_{0}, f'_{1,max}) $ values. It is clear that as we find the root of the straight line approximations for the solution $ \alpha_{*} $, the ranges of the solution $ \alpha_{*} \in [\alpha_{*,min}, \alpha_{*,max}]$ (in red crosses) is a lot larger for the bounded extrapolation than that for the interpolation. 

It is clear, that extrapolation may amplify variance of the optimal solution estimates. As a result, we give due consideration to whether our approximations interpolate or extrapolate in this study.




\section{Pseudo code and implementation details}



The general pseudo code is presented in Algorithm \ref{main_alg}.  The required inputs are listed, which include a flag for i) accepting $ flag = 1$ or ii) rejecting unbounded extrapolation updates $ flag = 0 $. The $ i $-th iteration is denoted by the subscript $ i $, and $ \hat{.} $ denotes the mini-batch sampling (\ref{mini-grad}). The line search algorithm starts with the inputs of the current mini-batch gradient $ \bm{\nabla}\hat{\bm{\mathcal{L}}}_{0,i} $ and search direction $ \boldsymbol{d}_{i} $.  The direction $ \boldsymbol{d} $ only needs to be a valid descent direction, but in this paper, we use the stochastic gradient descent (SGD) directions for simplicity, $ \boldsymbol{d}_{i} = -\bm{\nabla}\hat{\bm{\mathcal{L}}}_{0,i} $ at $ \alpha_{0} $. 

\begin{algorithm}[H]
	\label{main_alg}
	\DontPrintSemicolon 
	\KwIn{$ \bm{\nabla}\hat{\bm{\mathcal{L}}}_{0,i} $, $ \boldsymbol{d}_{i} $, $ \hat{f}_{0,i}$, $ flag $, $ \epsilon $, $ \alpha_{min} $, $ \alpha_{max} $}
	\KwOut{$ \alpha_{i} $,$ \bm{\nabla}\hat{\bm{\mathcal{L}}}_{0,i+1} $, $ \hat{f}_{0,i+1}$}
	Compute directional derivative $ \hat{f}'_{0,i} $ at $ \alpha_{0} $\;
	
	\If{$ |\hat{f}'_{0,i}|  < \epsilon$ \label{resampling}}
	{
		Recompute gradient $ \bm{\nabla}\hat{\bm{\mathcal{L}}}_{0,i+1} $ at $ \alpha_{0} $ with a new sub-sampled mini-batch for the next iteration.\; 
		\Return $ \alpha_{i} = 0 $, $\bm{\nabla}\hat{\bm{\mathcal{L}}}_{0,i+1}$
	}
	Initial guess $ \alpha_{1} = \|\boldsymbol{d}_{i}\|_{2}^{-1} $ \;
	\uIf{$ \alpha_{1} \geq \alpha_{max} $}{$\alpha_{1} = \alpha_{max}$}
	\ElseIf{$ \alpha_{1} \leq \alpha_{min} $}{$\alpha_{1} = \alpha_{min}$}
	\tcc{Option 1: Function-value-only approximation (f-f-f)}
	Compute $ \hat{f}_{1,i} $ at $ \alpha_{1} $ and $ \hat{f}_{2,i} $ at $ \alpha_{2} =  \alpha_{1}/2 $\;
	$\alpha_{*}  = \texttt{StepSizeFFF}(\alpha_{1},\alpha_{2},\hat{f}_{0,i},\hat{f}_{1,i},\hat{f}_{2,i}) $\;
	\tcc{Option 2: Mixed approximation (fg-f)}
	Compute $ \hat{f}_{1,i} $ at $ \alpha_{1} $\; 
	$\alpha_{*}  = \texttt{StepSizeFGF}(\alpha_{1},\hat{f}_{0,i},\hat{f}_{1,i},\hat{f}'_{0,i}) $;\tcp*{Appendix \ref{appendix}}
	\tcc{Option 3: Mixed approximation (f-fg)}
	Compute $ \hat{f}_{1,i}, \hat{f}'_{1,i} $ at $ \alpha_{1} $\;
	$\alpha_{*}  = \texttt{StepSizeFFG}(\alpha_{1},\hat{f}_{0,i},\hat{f}_{1,i},\hat{f}'_{1,i}) $;\tcp*{Appendix \ref{appendix}}
	\tcc{Option 4: Mixed approximation (fg-fg)}
	Compute $ \hat{f}_{1,i}, \hat{f}'_{1,i} $ at $ \alpha_{1} $\;
	$\alpha_{*}  = \texttt{StepSizeFGFG}(\alpha_{1},\hat{f}_{0,i},\hat{f}_{1,i},\hat{f}'_{0,i},\hat{f}'_{1,i}) $;\tcp*{Appendix \ref{appendix}}
	\tcc{Option 5: Derivative-only approximation (g-g)}
	Compute $ \hat{f}'_{1,i} $ at $ \alpha_{1} $\;
	$\alpha_{*}  = \texttt{StepSizeGG}(\alpha_{1},\alpha_{2},\hat{f}'_{0,i},\hat{f}'_{1,i}) $;\tcp*{Appendix \ref{appendix}}
	\uIf{$ (flag = 1) \And (\alpha_{*} \neq \alpha_{1}) \And (\alpha_{*}  > 0 )  $}{ \label{line_if}
		$\bm{\nabla}\hat{\bm{\mathcal{L}}}_{0,i+1} = \bm{\nabla}\hat{\bm{\mathcal{L}}}_{i}(\boldsymbol{x}_{0}+\alpha_{*}\boldsymbol{d}_{i})  $\;
	}
	\uElseIf{$ (flag = 0) \And (\alpha_{*} \neq \alpha_{1}) \And (0<\alpha_{*}<\alpha_{1})  $}{
		$\bm{\nabla}\hat{\bm{\mathcal{L}}}_{0,i+1} = \bm{\nabla}\hat{\bm{\mathcal{L}}}_{i}(\boldsymbol{x}_{0}+\alpha_{*}\boldsymbol{d}_{i})  $\;
		
	}\Else{$ \alpha_{*} = \alpha_{1} $}
	\Return{$ \alpha_{i} $,$ \bm{\nabla}\hat{\bm{\mathcal{L}}}_{0,i+1} $, $ \hat{f}_{0,i+1}$}
	\caption{Line search approximation}
\end{algorithm}


We first check whether the magnitude of the directional derivative $ \hat{f}'_{0,i} $ is greater than a prescribed tolerance of $ \epsilon $. We choose $ \epsilon $ to be as small as $ 10^{-16} $ because $ |\hat{f}'_{0,i}| $ needs to be greater than 0, but it still needs to be large enough not to cause any numerical errors. If it happens to be less than the tolerance, the local stochastic gradient $ \bm{\nabla}\hat{\bm{\mathcal{L}}}_{0,i} $ is recomputed using a new randomly sampled mini-batch to be returned for the next iteration with the final step size of $ \alpha_{i} = 0 $. 


If $ |\hat{f}'_{0,i}| $ is greater than $ \epsilon $, we can continue, and choose the initial step size $ \alpha_{1} $ based on the magnitude of the direction vector $ \boldsymbol{d} $, and we select it to be the inverse of the $ l_{2} $-norm of the search direction $ \|\boldsymbol{d}_{i}\|^{-1} $, as this provides small and exploiting step sizes for steeper directional derivatives $ |f'_{0}| >> 0 $ and large and exploring step sizes for flatter directional derivatives $ |f'_{0}| \approx 0 $. We then check if $ \alpha_{1} $ falls within the prescribed bounds denoted by $ \alpha_{min} $ and $ \alpha_{max}$, which are bounded to $ 10^{-7} $ and $ 10^{8} $, respectively, as used by \cite{Kafka2019}.  

Once, the initial guess $ \alpha_{1} $ is finalized, we can compute relevant information required to build an approximation of choice. Since we already have $ \hat{f}_{0,i} $ and $ \hat{f}'_{0,i} $ available, we need to additionally compute
\begin{enumerate}
	\item $ \hat{f}_{1,i} $ and $ \hat{f}_{2,i} $ for the f-f-f approximation,
	\item $ \hat{f}_{1,i} $ for the fg-f approximation,
	\item $ \hat{f}_{1,i} $ and $ \hat{f}'_{1,i} $ for the f-fg and fg-fg approximations,
	\item  $ \hat{f}'_{1,i} $ for the g-g approximation.
\end{enumerate}
The step sizes for each approximation are computed using \texttt{StepSizeFFF} (Algorithm~\ref{stepFFF}), \texttt{StepSizeFGF}, \texttt{StepSizeFFG}, \texttt{StepSizeFGFG} and \texttt{StepSizeGG} (Algorithms~\ref{stepFGF}-\ref{stepGG} in \ref{appendix}) for the f-f-f,  fg-f, f-fg, fg-fg and g-g approximations, respectively. 

\begin{algorithm}[H]
	\DontPrintSemicolon 
	\KwIn{$ \alpha_{1} $, $\alpha_{2} $, $ \hat{f}_{0} $, $ \hat{f}_{1} $, $ \hat{f}_{2} $, $ \epsilon_{k} $}
	\KwOut{$\alpha_{*}$}
	$ \alpha_{*} = \alpha_{1}$ \;
	Define a matrix $ \boldsymbol{A}_{1} $ and a vector $ \boldsymbol{b}_{1} $ from (\ref{fff})\;
	\If{$ rank(\boldsymbol{A}_{1}) = 3 $ \label{if_step1} }{
		
		Solve for the constants $	\boldsymbol{k} = \boldsymbol{A}_{1}^{-1}\boldsymbol{b}_{1} $ from (\ref{quad_func})\;
		\If{$ k_{1}  > \epsilon_{k} $ \label{if_step2} }{
			$\alpha_{*} =  -k_{2}/(2k_{1}) $\;
			\uIf{$ \alpha_{*} \geq \alpha_{max} $}{$\alpha_{*} = \alpha_{max}$}
			\ElseIf{$ \alpha_{*} \leq \alpha_{min} $}{$\alpha_{*} = \alpha_{min}$}
		}
	}
	\caption{\texttt{StepSizeFFF}}
	\label{stepFFF}
\end{algorithm}

The respective step size functions ensure that the matrix $ \boldsymbol{A} $ is not singular (e.g. line~\ref{if_step1} in Algorithm~\ref{stepFFF}). We check whether the curvature $ k_{1} $ is a positive value or greater than a prescribed tolerance $ \epsilon_{k} $, which we set to be  $ 10^{-18} $. This ensures that the approximation is convex, and that it does not run into numerical issues for computing the step size $ \alpha_{*} $ (e.g. line~\ref{if_step2} in Algorithm~\ref{stepFFF}). Again, we check if the computed step size is within the upper and lower bounds ($ \alpha_{max} $ and $ \alpha_{min} $). Note that the linear system of the fg-fg approximation (\ref{fgfg}) is solved using the least square regression, and that when implementing the g-g case, the rank of the matrix $ \boldsymbol{A}_{5} $ (\ref{gg}) needs to be 2 since there are only 2 constants (i.e. $ k_{1} $ and $ k_{2} $) that are solved.

Table~\ref{table:1} shows that the number of function evaluations required may vary between the types of approximations, and the choices made in an iteration. The different choices are 1) resample when the initial magnitude of the directional derivative is too small (line~\ref{resampling} in Algorithm~\ref{main_alg}), 2) the immediate accept condition (I.A.C.) if the approximation is concave and 3) interpolation or bounded extrapolation when the approximation is convex. Note we count one forward and backward process as one function evaluations following \cite{Mahsereci2017a}. Hence, the f-f-f approximation, which requires function evaluations at 3 points, is more computationally expensive than the other approximation. Note that with-extrapolation and without-extrapolation, Algorithm~\ref{main_alg} requires same number of function evaluations for both, as a function evaluation at $ \alpha_{1} $ is required to identify whether $ \alpha_{*} $ is computed using interpolation or bounded extrapolation.

\begin{table}
	\tbl{Number of function evaluations requires for each type of approximation for 1)~resampling, 2)~immediate accept condition (I.A.C.) for a concave approximation and 3)~interpolation or extrapolation a convex approximation.}
	{\begin{tabular}{lccc}
		\toprule
		                        & \multicolumn{3}{c}{Function evaluations}          \\
		\cmidrule{2-4}
			Approx. & Resample &        I.A.C.         & Interp./Extrap.   \\ \midrule
		f-f-f                   & 1  &         2         &  3     \\
		fg-f                    & 1  &         1         &  2     \\
		f-fg                     & 1  &         1         &  2     \\
		fg-fg                    & 1  &         1         &  2     \\
		g-g                   & 1  &         1         &  2     \\ \bottomrule
	\end{tabular}}
	\label{table:1}
\end{table}

\section{Preliminary numerical  study}
In this section, we conduct 2 preliminary numerical experiments; Firstly, we discuss the risk of accepting bounded extrapolation estimates by comparing with-extrapolation line searches against without-extrapolation line searches. The aim is to understand the variance of the extrapolation estimates and the risk when these accepting estimates when training neural network models. Secondly, we investigate the characteristics of the 5 approximations, (\ref{fff})-(\ref{gg}) constructed using different pieces of information. The aim of this is to observe the variance of the solutions $ \alpha_{*} $ resulted from each approximation, and quantify the characteristics of the solutions.

The test problem used for the two studies is a zero hidden layer, single output logistic network for the breast cancer dataset (400 train samples and 169 test samples) with the sigmoid activation function and binary cross entropy loss function. The starting guesses for weights $ \boldsymbol{x} $ are generated from a standard normal distribution and the training continues for $ 10^{5} $ function evaluations.  The minimum $ \alpha_{min} $ and maximum $ \alpha_{max} $ step sizes chosen are $ 10^{-8} $ and $ 10^{7} $, respectively. 

\subsection{Risk of accepting bounded extrapolation results}
\begin{figure}
	\centering
	\includegraphics[width=1.0\linewidth]{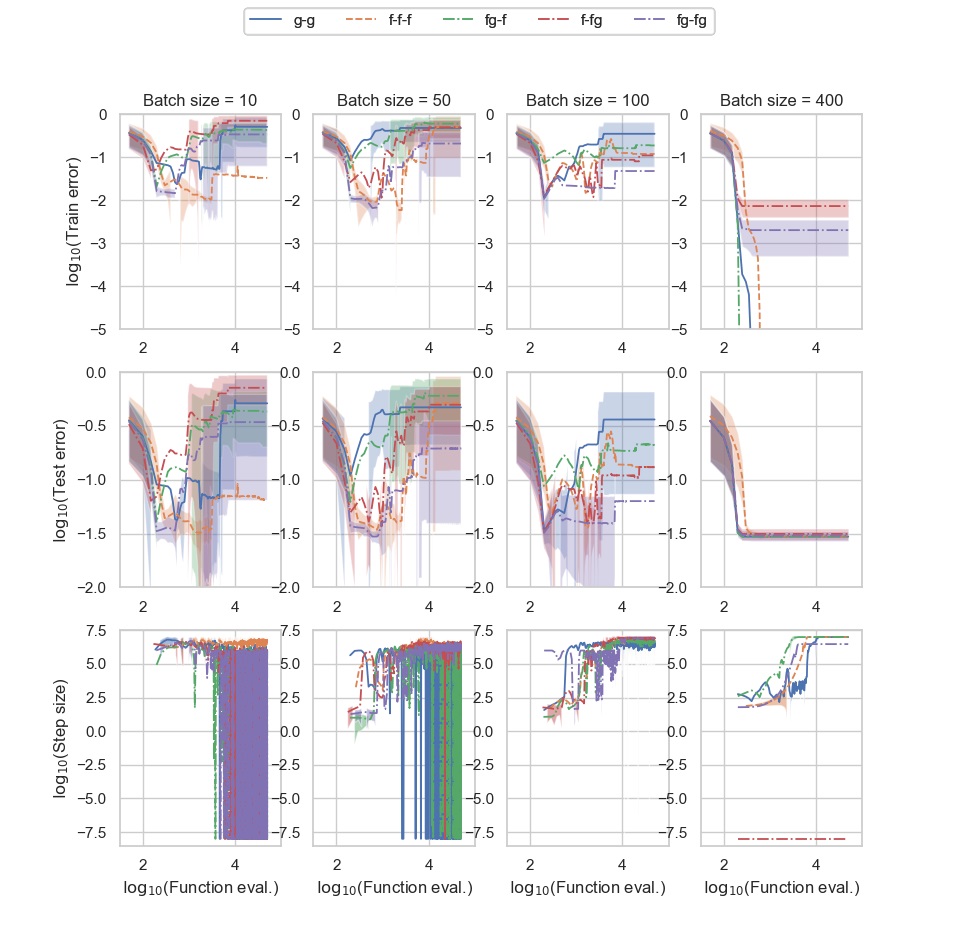}
	\caption{Bounded extrapolation allowed: classification train errors and test errors in $ log_{10} $ scale for 1) derivative-only (g-g) approximation 2) function-value-only (f-f-f) approximation 3) mixed approximation with directional derivative only at $ \alpha_{0} $ (fg-f) 4) mixed approximation with directional derivative only at $ \alpha_{1} $ (f-fg) and 5) mixed approximation with directional derivative information at both $ \alpha_{0} $ and $ \alpha_{1} $ (fg-fg) approximation-assisted line search for different batch sizes (10, 50, 100 and 400) for the breast cancer dataset.}
	\label{cancer1}
\end{figure}

\begin{figure}
	\centering
	\includegraphics[width=1.0\linewidth]{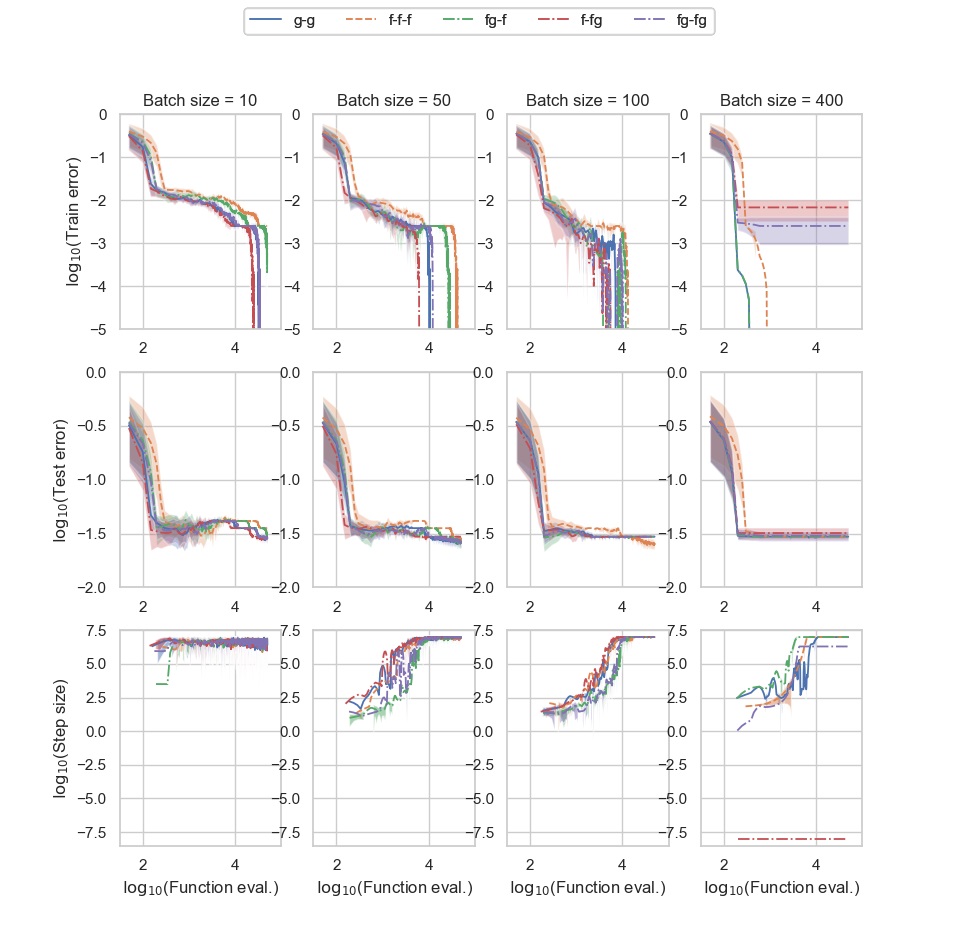}
	\caption{Bounded extrapolation disallowed: classification train errors and test errors in $ log_{10} $ scale for 1) derivative-only (g-g) approximation 2) function-value-only (f-f-f) approximation 3) mixed approximation with directional derivative only at $ \alpha_{0} $ (fg-f) 4) mixed approximation with directional derivative only at $ \alpha_{1} $ (f-fg) and 5) mixed approximation with directional derivative information at both $ \alpha_{0} $ and $ \alpha_{1} $ (fg-fg) approximation-assisted line search for different batch sizes (10, 50, 100 and 400) for the breast cancer dataset.}
	\label{cancer2}
\end{figure}

The results for the five 1D quadratic approximations on the logistics network with and without bounded extrapolations are shown in Figures~\ref{cancer1} and \ref{cancer2}, respectively. The results depicted the mean and standard deviation for 10 runs. The $ \log_{10} $ scaled results are shown for training error, testing error and step sizes, each as a row. Along the  columns, the batch size, $ m $ varies; 10, 50, 100 and 400. The with-extrapolation results in Figure~\ref{cancer1} show that the errors of all the approximations seem to converge for about 100 function evaluations, but quickly diverge, and the variance of the step sizes is larger for the smaller batch sizes, $ m $. For the full batch $ m = 400 $, the fg-f, g-g and f-f-f approximation perform well as the loss function becomes continuous without sampling errors.

The without-extrapolation results in Figure~\ref{cancer2} show that, overall, both the training and testing errors are significantly lower when compared to the with-extrapolation results. In addition, the variance of the step size is also smaller when disallowing bounded extrapolation estimates. For the larger batch size, the convergence rate of the train error is faster. The perturbation of the test error decreases as the $ m $ increases, but the approximation certainly finds lower minima when using mini-batch \cite{Masters2018}. Note that the results of the full batch in Figure~\ref{cancer2} is similar to that of Figure~\ref{cancer1}. Whereas the g-g and fg-f perform almost the same, and outperform the other approximations, the f-fg approximation results indicate hampered improvement due to small step sizes. The fg-fg approximation, which uses the most amount of information, finds poorer minima, although its step size is similar to the best performing step sizes.

In conclusion, disallowing bounded extrapolation helps to decrease the errors, reduces variance, and ensure more stable training. Hence, we implement the line search by disallowing bounded extrapolation for the remainder of the study. The results clearly show that approximations constructed using fewer information can outperform approximations constructed using more information. This clearly indicates the importance of selecting appropriate information.

\subsection{Characteristics of the various approximations}
\begin{figure}
	\centering
	\subfloat[Function-value-only (f-f-f)]{%
		\resizebox*{4cm}{!}{\includegraphics{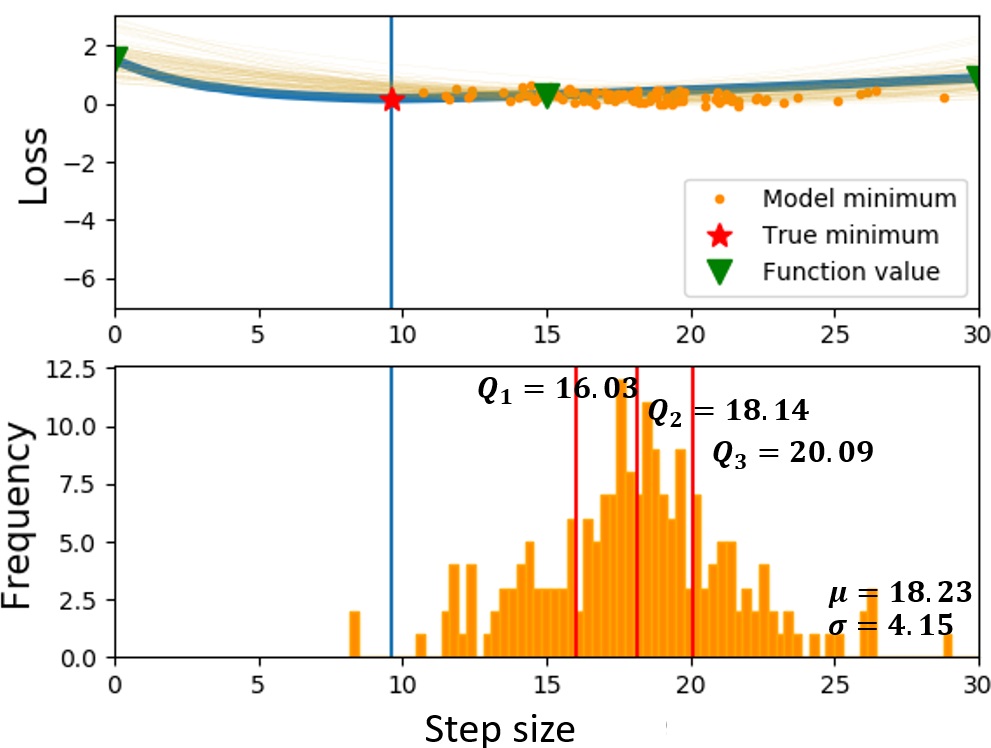}}}\hspace{5pt}
	\subfloat[Mixed (fg-f)]{%
		\resizebox*{4cm}{!}{\includegraphics{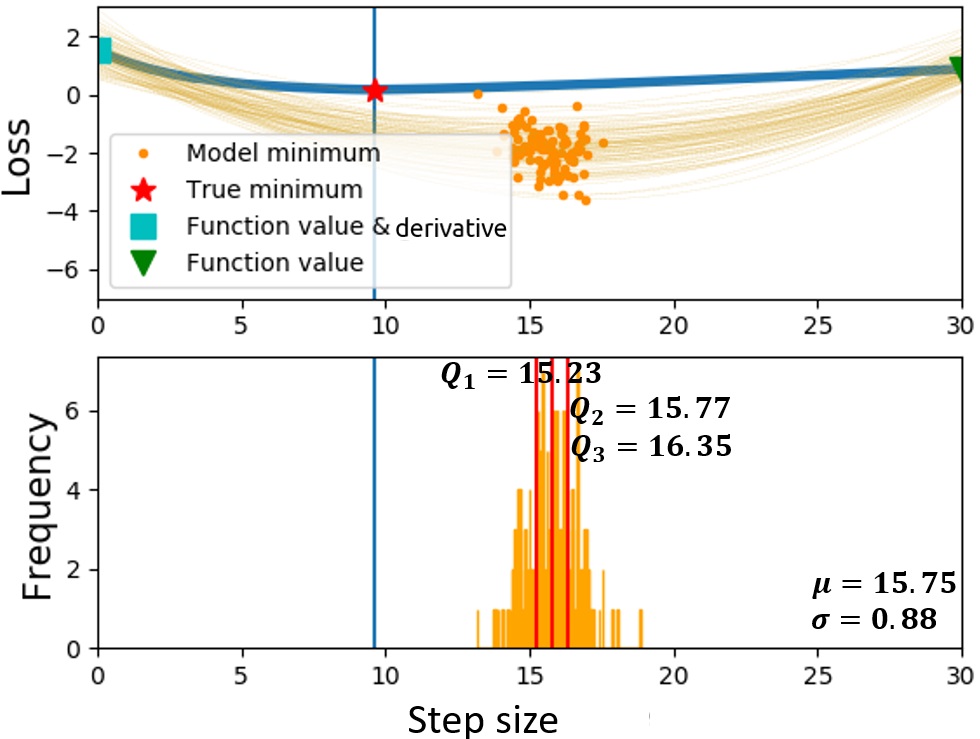}}} \hspace{5pt}
	\subfloat[Mixed (f-fg)]{%
	\resizebox*{4cm}{!}{\includegraphics{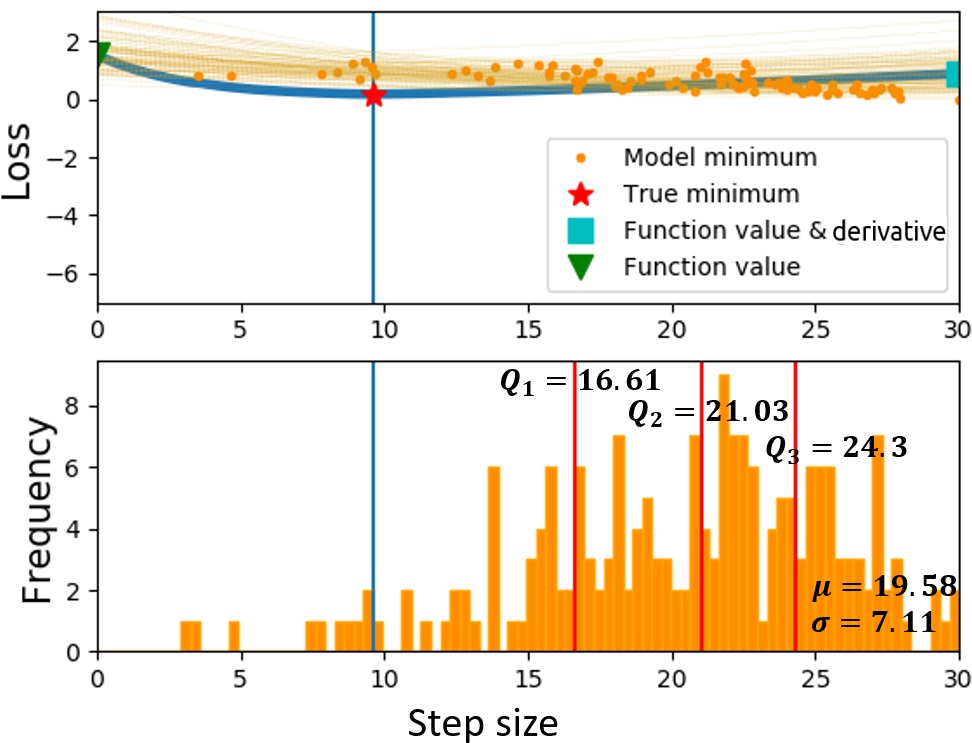}}}\hspace{5pt}
	\subfloat[Mixed (fg-fg)]{%
	\resizebox*{4cm}{!}{\includegraphics{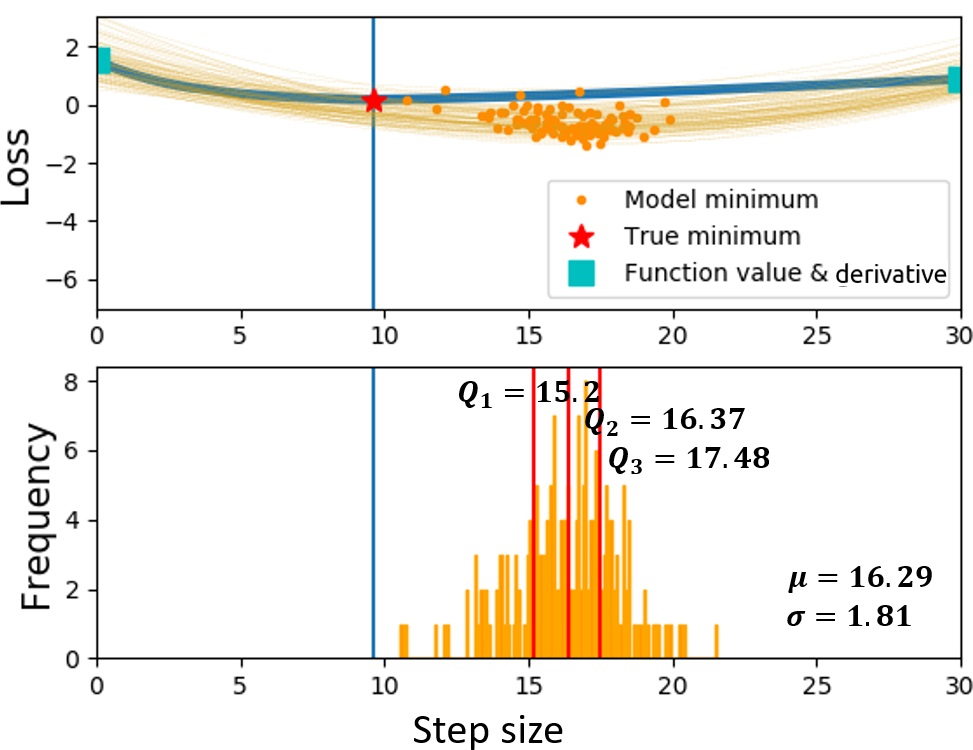}}}\hspace{5pt}
	\subfloat[Derivative-only (g-g)]{%
	\resizebox*{4cm}{!}{\includegraphics{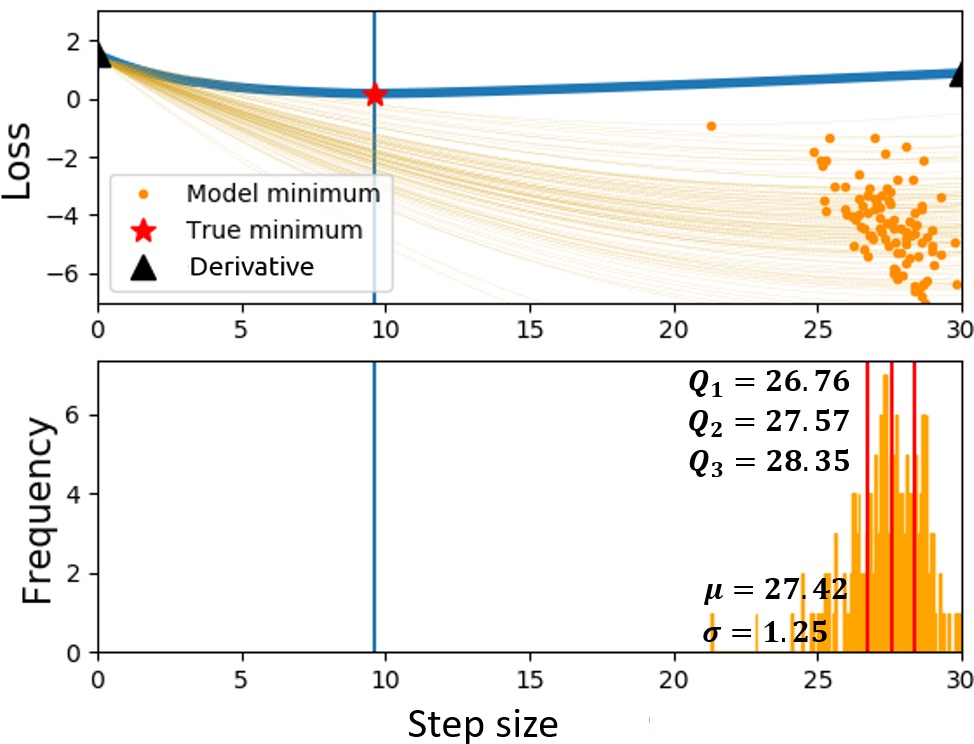}}}
	\caption{Illustration of quadratic approximations built using different information; (a) function-value-only (f-f-f), (b) mixed (fg-f), (c) mixed (f-fg), (d) mixed (fg-fg), (e) derivative-only (g-g). The frequency of 200 approximation-minima are shown in the histogram (bottom), and only randomly selected 100 approximations are shown in loss domain (top). The derivative-only approximations are plotted on the loss domain, not directional derivative domain by adding arbitrary constants.} \label{surrogate}
\end{figure}

The fact that approximations using less but selective information outperform approximations using more information, inspired this investigation to quantify the distribution of the estimates for the five approximations. Shown in Figure~\ref{surrogate} for the breast cancer logistic regression task are the distributions of the solution estimates of 200 fitted quadratic approximations. Each approximation is constructed using dynamic MBSS with 50 samples per mini-batch. Note that for each plot, we only show a random selection of 100 approximations above in the interest of visual clarity, but show the distribution of solutions from all 200 approximations below. In order for direct comparison of all approximations in the same domain of loss, we converted the derivative-only approximation (g-g) of derivative domain into function value approximations by integrating it with respect to the step size $ \alpha $ and added arbitrary constants $ k_{3} $ in (\ref{quad_func}), which is independent of step sizes. 

Figure~\ref{surrogate} also shows the mean $ \mu $, standard deviation $ \sigma $, 1st, 2nd (median) and 3rd quartiles ($ Q_{1}, Q_{2} $ and $ Q_{3} $) for each distribution. The minimum of the true full-batch loss is at $ \alpha = 9.6 $. The mean solutions $ \mu $ of all distributions are commonly far from the true minimum while the derivative-only approximation is the furthest. As previously discussed, the bias (difference between the mean and actual solution) of the approximations reflects our choice of most basic quadratic approximation, which can be improved by selecting a more flexible approximation, that would improve the overshooting phenomena outlined in Figure~\ref{shooting}. 

Conversely, the variance is more indicative of the quality of information to which we will give due consideration when interpreting the results. The approximations with the least variance, i.e. fg-f and g-g, in the estimate, consistently outperformed the other approximations in the previous section as shown in Figures~\ref{cancer1} and \ref{cancer2}. It is noted that the approximations with  derivative information at the starting point $ \alpha_{0}  $ commonly have lower standard deviation $ \sigma $ compared to approximations, that do not enforce derivatives at $ \alpha_{0} $. These preliminary studies demonstrate the importance of being selective in choosing information to construct approximations, as more information, e.g. fg-fg approximation, may severely hamper performance compared to the fg-f and g-g approximations.

\section{Experimental setup}
In this section, we consider a comprehensive numerical investigation to our preliminary investigation to identify aspects in our preliminary study, that generalize well over larger and more complex neural network models.  We consider the experimental study as outlined by \cite{Mahsereci2017a}, which include (1) MNIST \cite{LeCun1998} and (2) CIFAR-10 \cite{Krizhevsky2009a}. The characteristics of the datasets and the experimental settings for each dataset are given as follows:

\begin{itemize} 	
	\item MNIST is a 10-class classification dataset with  $ 28\times28 $ input size, $ 6\times10^{4} $ train samples and $1\times10^{4} $ test samples. The batch sizes chosen for MNIST are 10, 100, 200 and 1000. The maximum number of function evaluations was $ 4\times10^{4} $.
	\item CIFAR-10 is a 10-class classification dataset with  $ 32\times32 $ input size, $ 5\times10^{4} $ train samples and $1\times10^{4} $ test samples. Note that we used the ``batch 1".  The batch sizes chosen for CIFAR-10 are 10, 100, 200 and 1000. The maximum number of function evaluations was $ 10^{4} $.
\end{itemize}

Both MNIST and CIFAR-10 datasets were trained on two fully connected feedforward nets with biases; N-I and N-II. 
\begin{itemize}
	\item N-I: it has a shallow network structure of a single hidden layer; $ n_{input}$-800-$n_{ouput}$ with sigmoid activation function and a cross entropy loss. The initial weights are sampled from a normal distribution.
	\item N-II: it has a deep network structure of three hidden layers; $ n_{input}$-1000-500-250-$n_{output}$ with tanh activation function and a square loss. The initial weights are obtained using the Xavier initialization \cite{glorot2010understanding}.
\end{itemize}
Note that all the problems are one-hot encoded. We will investigate the line search methods on the 4 problems (i.e. MNIST with N-I, N-II and  CIFAR-10 with N-1, N-II) under 3 conditions. Note also that the line search methods (Algorithm~\ref{main_alg}) omit the bounded extrapolation update, and takes the immediate accept solution, which is $ \alpha_{*} = \alpha_{1} $. The search direction is set to be the stochastic gradient descent (SGD) direction.

The three conditions are, namely, 1) no step size limit test, 2) enforced step size limit test and 3) exact line search test. For the no step size limit investigation, we omit checking step size limits with $ \alpha_{max} $ and $ \alpha_{min} $. Hence, we accept every estimate the approximation produces as the step size, whether it extrapolates or not. The aim of this investigation is to identify the stability and robustness of each approximation in an uncontrolled environment. 

For the enforced step size limit test, we simply test Algorithm~\ref{main_alg} on the four problems. The aim of this is to test the five approximations on more complex problems compared to the breast cancer dataset task, and also observe the performance compared to the no limit cases. Note that both no step size limit test and enforced step size limit test conduct the training ten times at different starting points to compute the mean and standard deviation.

Lastly, for the exact line search test, we test the line search algorithm using a larger fixed batch of $ 10^{4} $ samples, and compare the performance of the approximations to the exact line search using the golden-section method \cite{arora2004introduction}. The aim of this test is to investigate the step sizes and the convergence rates of the line search using the approximations compared to those of the exact line search with respect to the number of iterations. Note that using a fixed batch allows for a deterministic optimization. Hence, we only conduct one run of 3000 iterations.

\section{Experimental results}
\subsection{No upper and lower bound limits}
\begin{figure}
	\centering
	\includegraphics[width=1\linewidth]{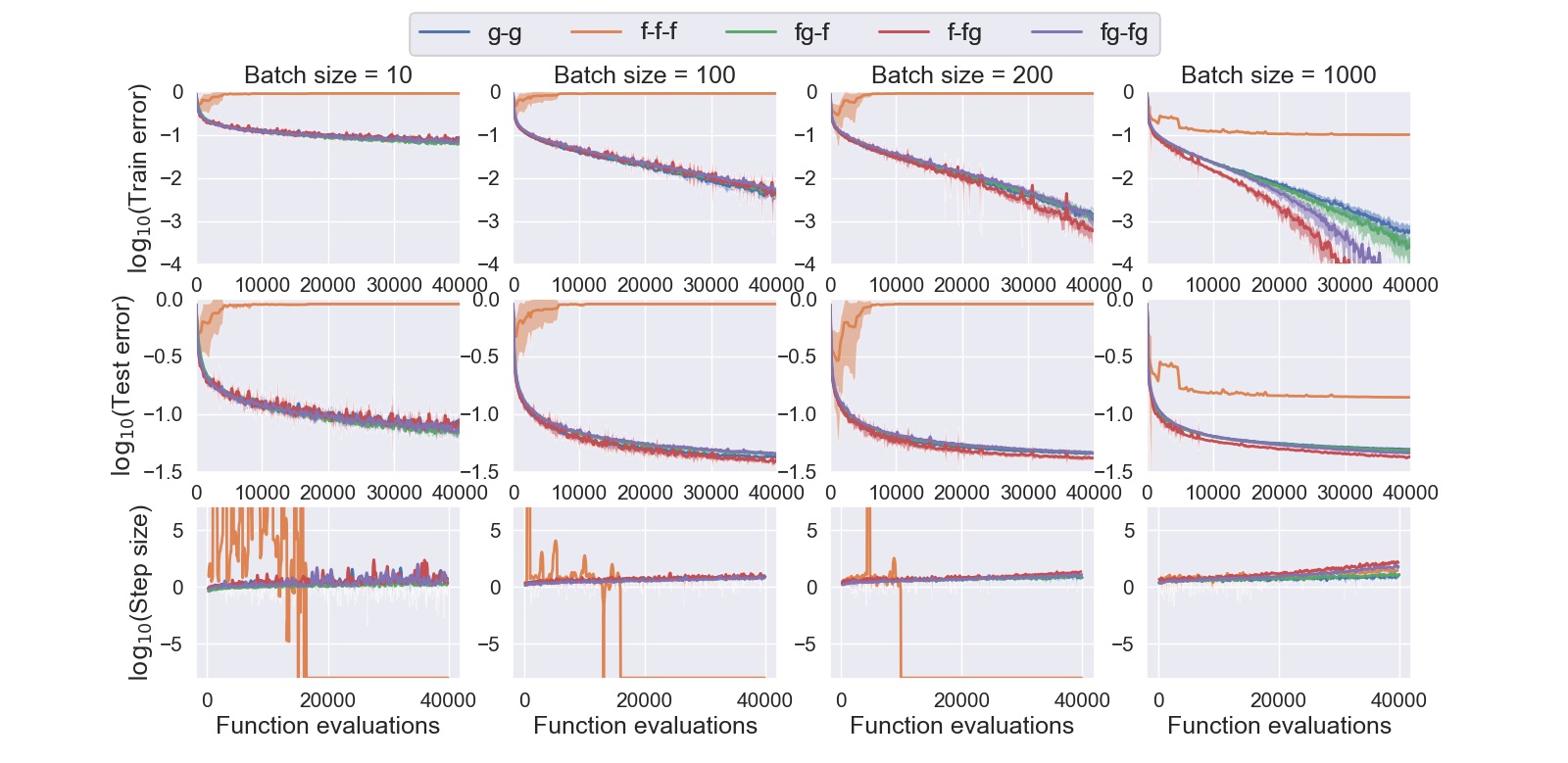}
	\caption{
		No bound limits: comparison of approximation-assisted line search methods with various information; 1) derivative-only (g-g), 2) function-value-only (f-f-f), 3) mixed approximation with function values at both $ \alpha_{0} $ and $ \alpha_{1} $ and directional derivative only at $ \alpha_{0} $ (fg-f), 4) directional derivative only at $ \alpha_{1} $ (f-fg) and 5) directional derivatives at both $ \alpha_{0} $ and $ \alpha_{1} $ (fg-fg)  on MNIST dataset with N-I architecture with respect to function evaluations.
	}\label{prob_d1n1_0}
\end{figure}

\begin{figure}
	\centering
	\includegraphics[width=1\linewidth]{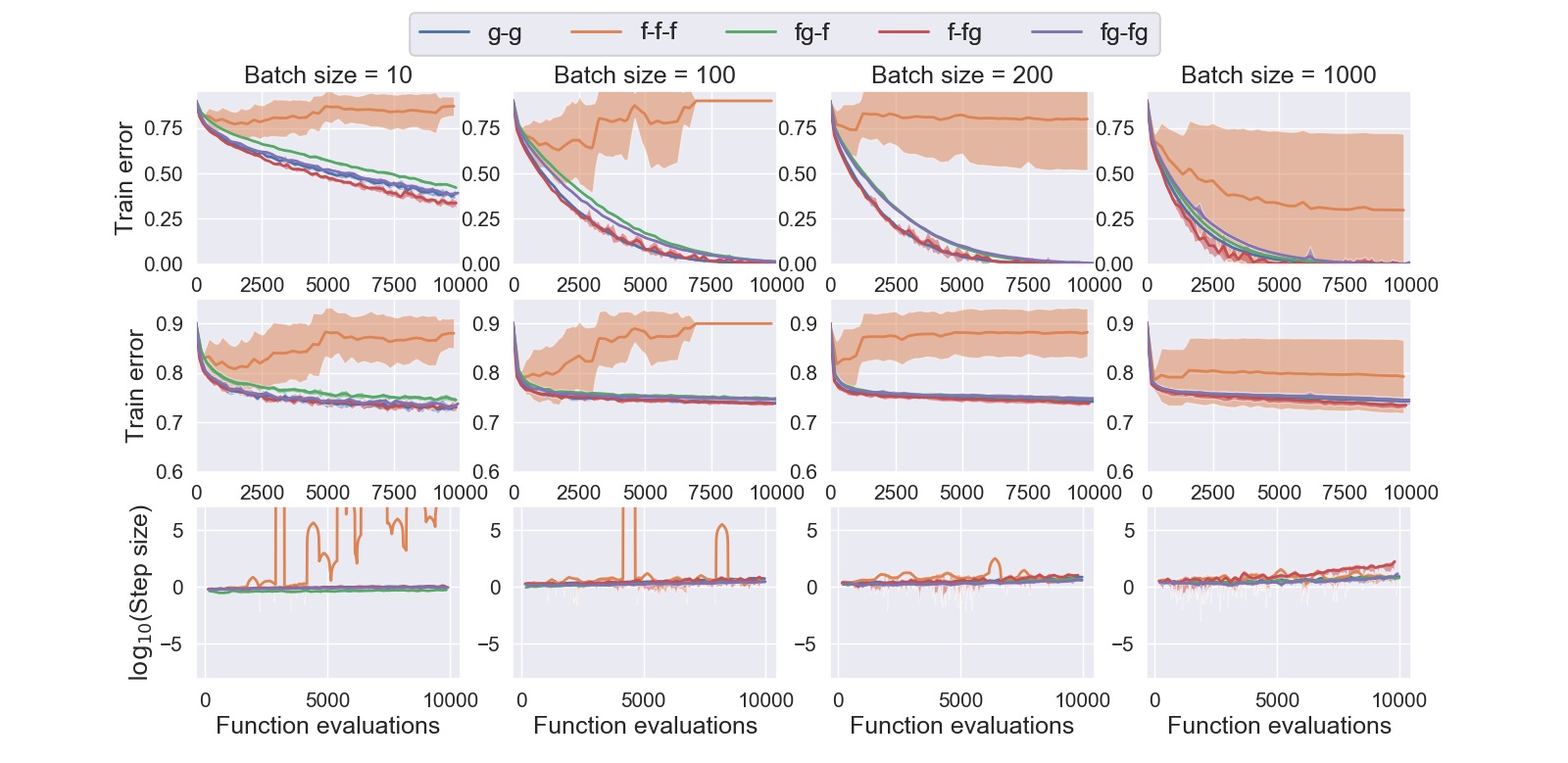}
	\caption{
		No bound limits: comparison of approximation-assisted line search methods with various information; 1) derivative-only (g-g), 2) function-value-only (f-f-f), 3) mixed approximation with function values at both $ \alpha_{0} $ and $ \alpha_{1} $ and directional derivative only at $ \alpha_{0} $ (fg-f), 4) directional derivative only at $ \alpha_{1} $ (f-fg) and 5) directional derivatives at both $ \alpha_{0} $ and $ \alpha_{1} $ (fg-fg)  on CIFAR-10 dataset with N-I architecture with respect to function evaluations.
	}\label{prob_d2n1_0}
\end{figure}

\begin{figure}
	\centering
	\includegraphics[width=1\linewidth]{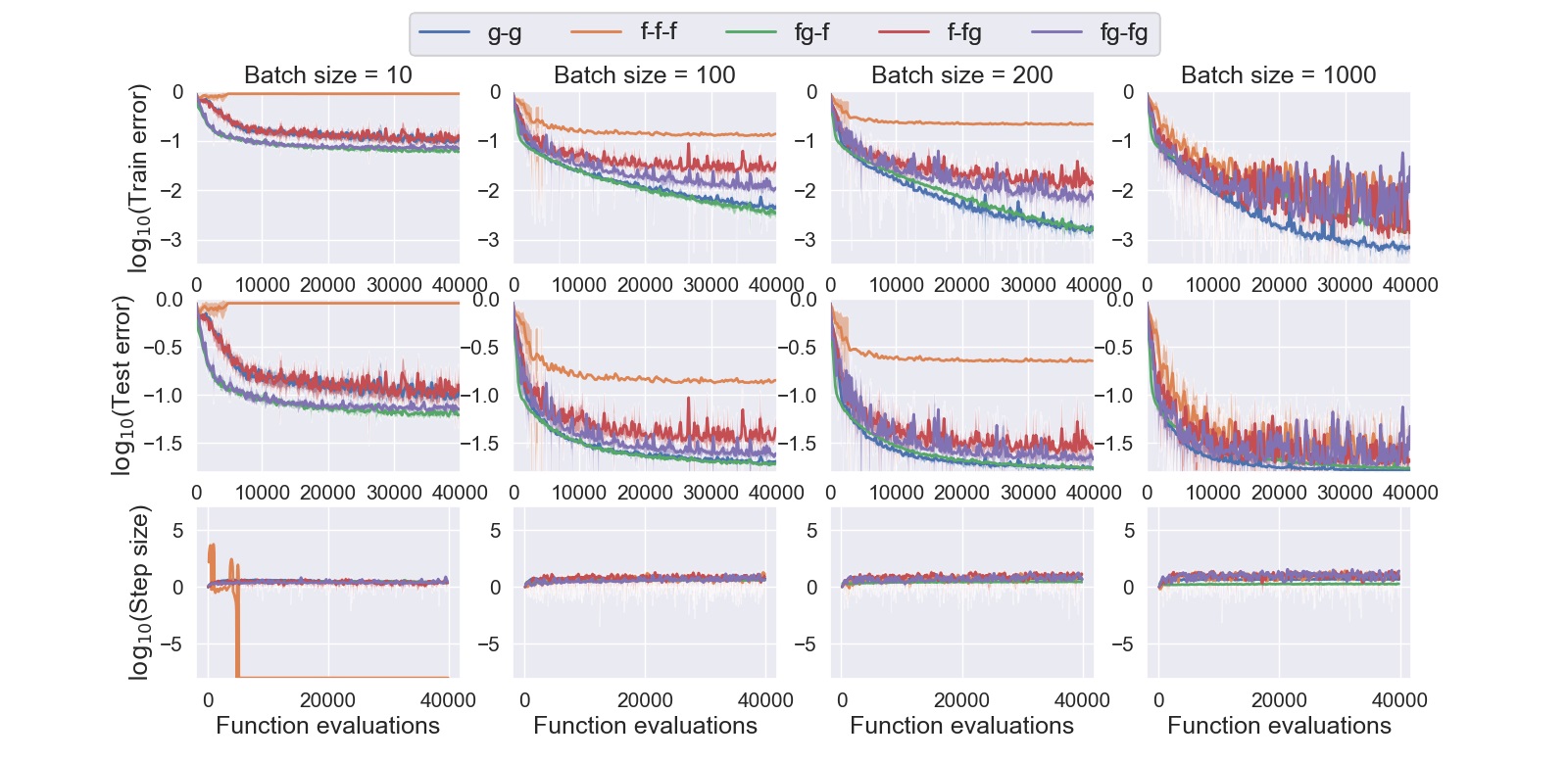}
	\caption{
		No bound limits: comparison of approximation-assisted line search methods with various information; 1) derivative-only (g-g), 2) function-value-only (f-f-f), 3) mixed approximation with function values at both $ \alpha_{0} $ and $ \alpha_{1} $ and directional derivative only at $ \alpha_{0} $ (fg-f), 4) directional derivative only at $ \alpha_{1} $ (f-fg) and 5) directional derivatives at both $ \alpha_{0} $ and $ \alpha_{1} $ (fg-fg)  on MNIST dataset with N-II architecture with respect to function evaluations.
	}\label{prob_d1n2_0}
\end{figure}

\begin{figure}
	\centering
	\includegraphics[width=1\linewidth]{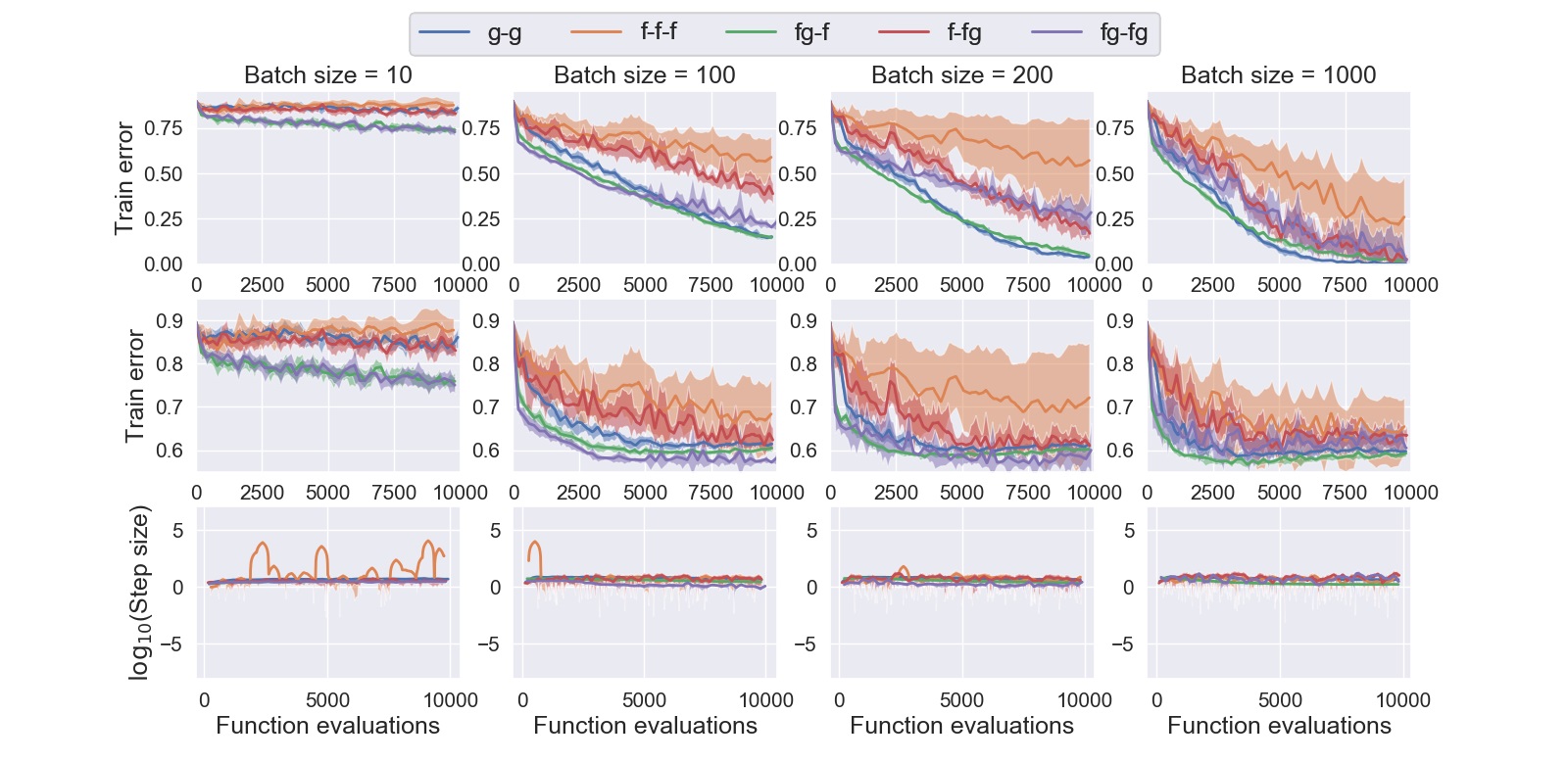}
	\caption{
		No bound limits: comparison of approximation-assisted line search methods with various information; 1) derivative-only (g-g), 2) function-value-only (f-f-f), 3) mixed approximation with function values at both $ \alpha_{0} $ and $ \alpha_{1} $ and directional derivative only at $ \alpha_{0} $ (fg-f), 4) directional derivative only at $ \alpha_{1} $ (f-fg) and 5) directional derivatives at both $ \alpha_{0} $ and $ \alpha_{1} $ (fg-fg)  on CIFAR-10 dataset with N-II architecture with respect to function evaluations.
	}\label{prob_d2n2_0}
\end{figure}

The training results of the no step size limit test for N-I with MNIST, CIFAR-10 are shown in Figures~\ref{prob_d1n1_0} and \ref{prob_d2n1_0}. The results for N-II with both datasets are shown in Figures~\ref{prob_d1n2_0} and \ref{prob_d2n2_0}. The first row shows train error, the second row shows the test error and the last row shows the step sizes in the log scale. Note that the errors for the MNIST datasets are in the log scale. In columns, we increase the batch sizes from left to right. Although there are no bounds specified for the step sizes, it is only the f-f-f approximation, which provide exploding step sizes as results, and the approximations with at least one piece of directional derivative information, show stable results overall. 

\begin{figure}
	\centering
	\includegraphics[width=0.5\linewidth]{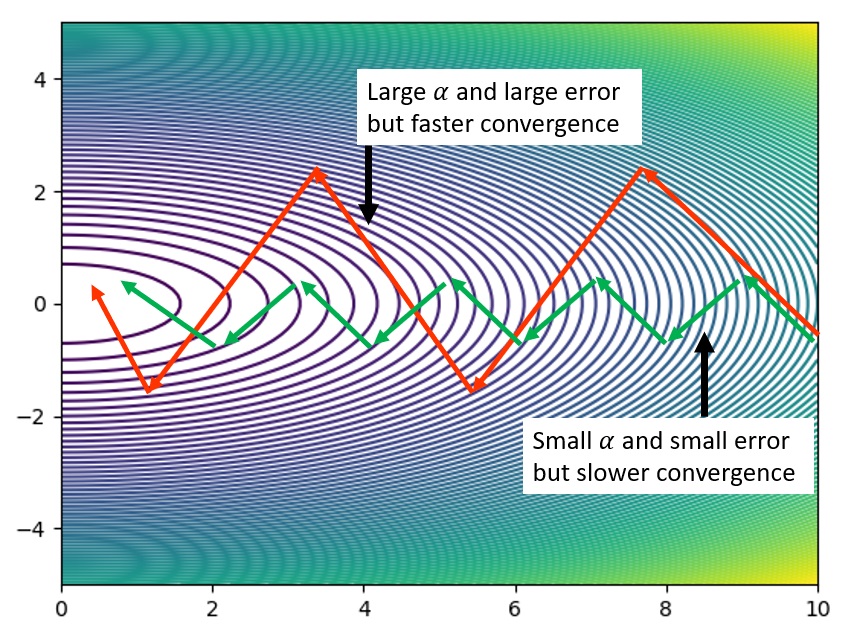}
	\caption{Illustration of the improvements by made the larger (in red) and smaller (in green) step sizes along the noisy descent directions: the larger step sizes $ \alpha $ cause larger perturbation in the errors by traveling along the steeper contour, but it makes faster convergence compared to the smaller and more exact step sizes traveling along the flatter contour.}
	\label{contour}
\end{figure}

As we can observe from Figures~\ref{prob_d1n1_0} and \ref{prob_d2n1_0}, the larger the sizes of mini-batches become, the lower the training errors get, and the convergence rates also visibly increase. Because we know that both the f-fg and f-f-f approximations had large variances in solutions as shown in Figure~\ref{surrogate}, we might expect both approximations underperform compared to the less variance approximations but the f-fg approximation actually outperformed all the rest, and this is more evident when the batch size is 1000. This result was obtained because the average step sizes of the f-fg was relatively larger than the other approximations. As shown in Figure~\ref{contour},  approximations, which produce larger step sizes, tend to have the faster convergence rate, because the larger step sizes indicated in red tend to travel faster along the true descent direction, often by overshooting along slanted high curvatures directions compared to more exact but smaller step sizes indicated in green. The identical phenomenon is illustrated in the N-I architecture for both MNIST and CIFAR-10.

No bound limit results for the N-II architecture on MNIST and CIFAR-10 datasets are shown in Figures~\ref{prob_d1n2_0} and \ref{prob_d2n2_0}. For small batch sizes $ m=10 $, both fg-f and fg-fg, which had small variance in the distribution of solutions in Figure~\ref{surrogate}, outperform. This shows that when the information is strongly biased due to small $ m $, it is better to have even a little more information such as the fg-fg approximation. However, as $ m $ increases, both the fg-f and g-g approximation start outperform the rest and when $ m = 1000 $, the g-g approximation finds the lowest errors. The results demonstrate that the approximations with less, but only relevant information outperform the approximations with more but irrelevant information. 

\subsection{Enforced upper and lower bound limits}

\begin{figure}
	\centering
	\includegraphics[width=1\linewidth]{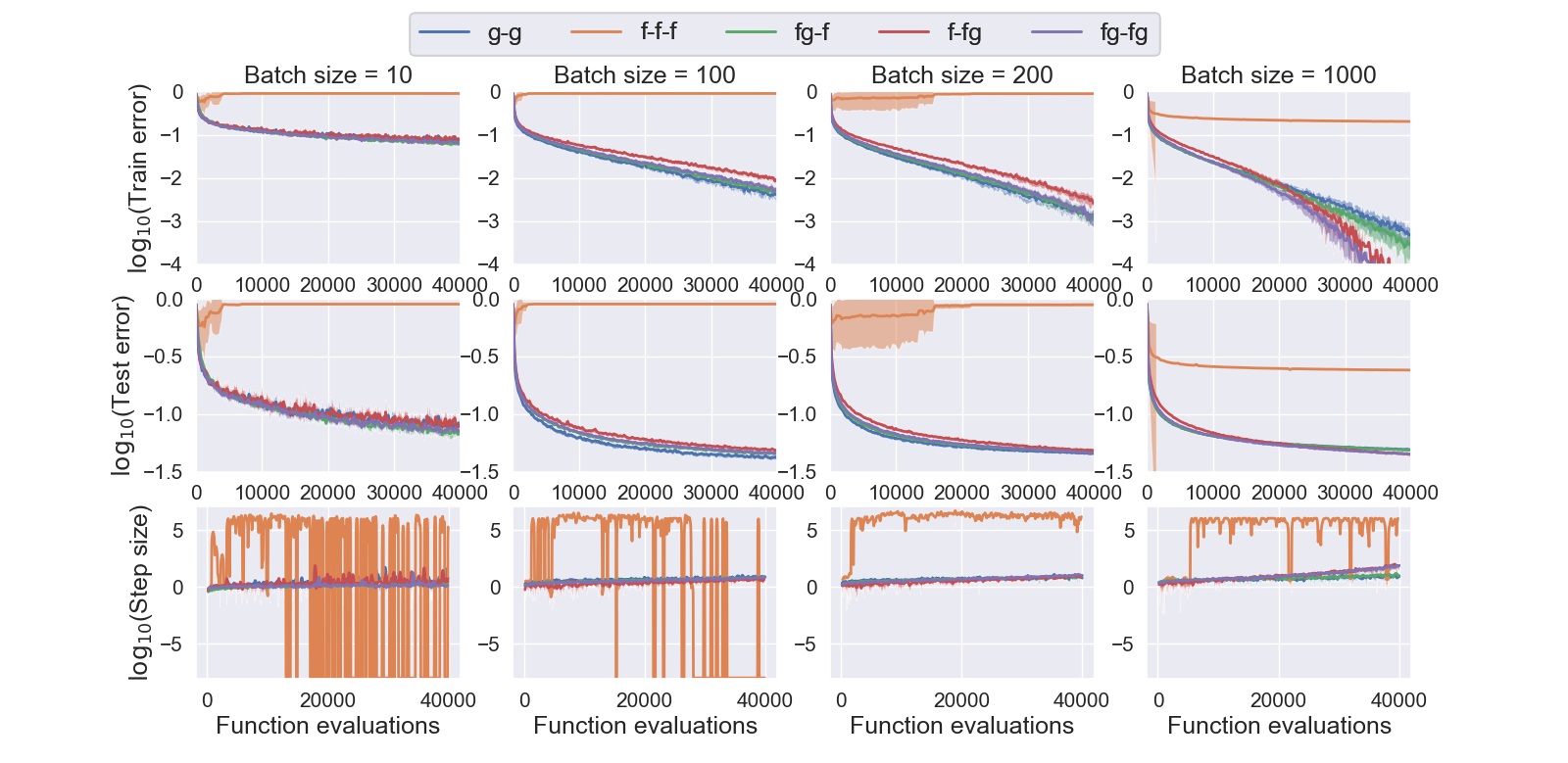}
	\caption{
		Enforced bound limits: comparison of approximation-assisted line search methods with various information; 1) derivative-only (g-g), 2) function-value-only (f-f-f), 3) mixed approximation with function values at both $ \alpha_{0} $ and $ \alpha_{1} $ and directional derivative only at $ \alpha_{0} $ (fg-f), 4) directional derivative only at $ \alpha_{1} $ (f-fg) and 5) directional derivatives at both $ \alpha_{0} $ and $ \alpha_{1} $ (fg-fg)  on MNIST dataset with N-I architecture with respect to function evaluations.
	}\label{prob_d1n1_1}
\end{figure}

\begin{figure}
	\centering
	\includegraphics[width=1\linewidth]{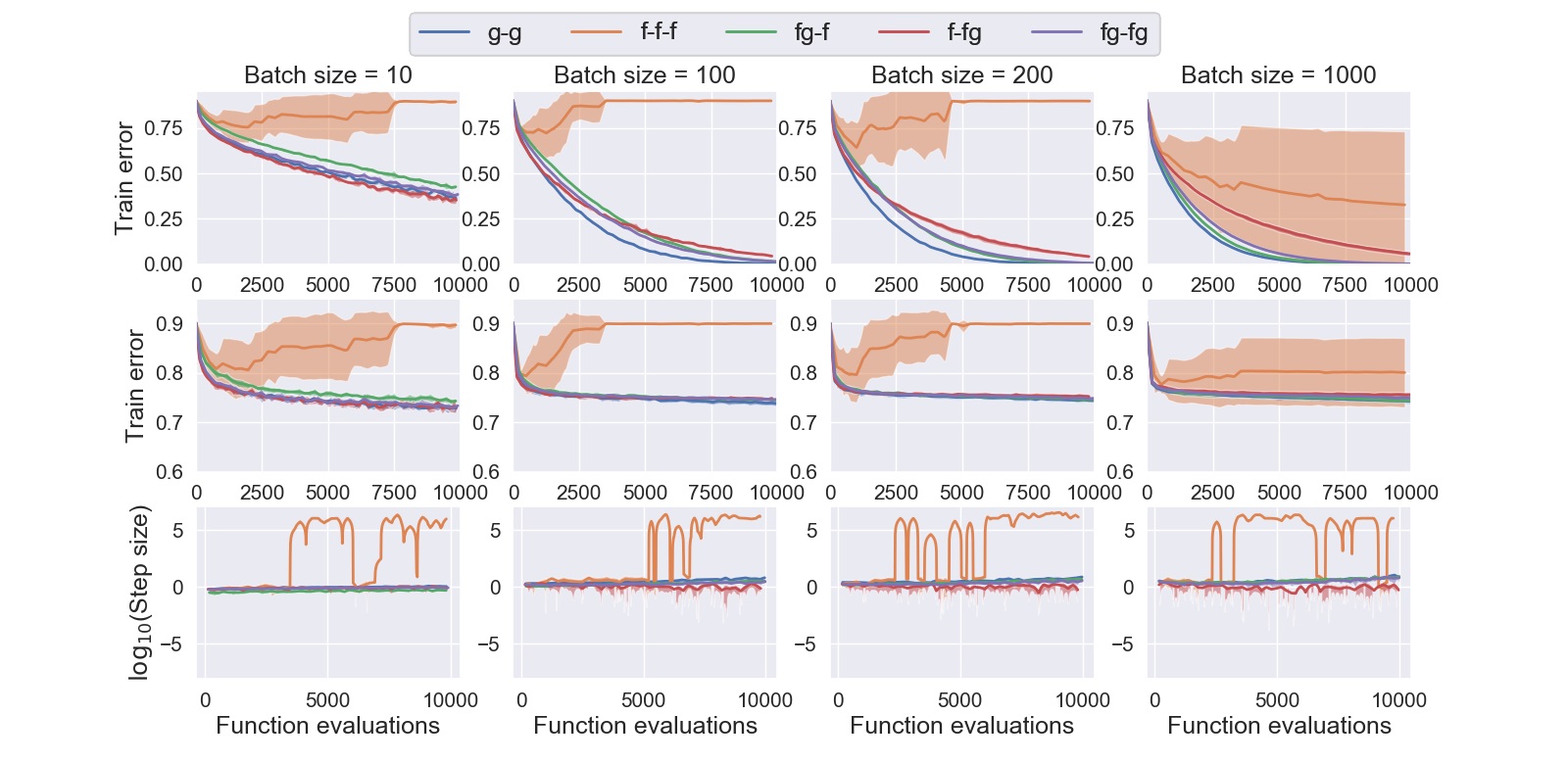}
	\caption{
		Enforced bound limits: comparison of approximation-assisted line search methods with various information; 1) derivative-only (g-g), 2) function-value-only (f-f-f), 3) mixed approximation with function values at both $ \alpha_{0} $ and $ \alpha_{1} $ and directional derivative only at $ \alpha_{0} $ (fg-f), 4) directional derivative only at $ \alpha_{1} $ (f-fg) and 5) directional derivatives at both $ \alpha_{0} $ and $ \alpha_{1} $ (fg-fg)  on CIFAR-10 dataset with N-I architecture with respect to function evaluations.
	}\label{prob_d2n1_1}
\end{figure}

\begin{figure}
	\centering
	\includegraphics[width=1\linewidth]{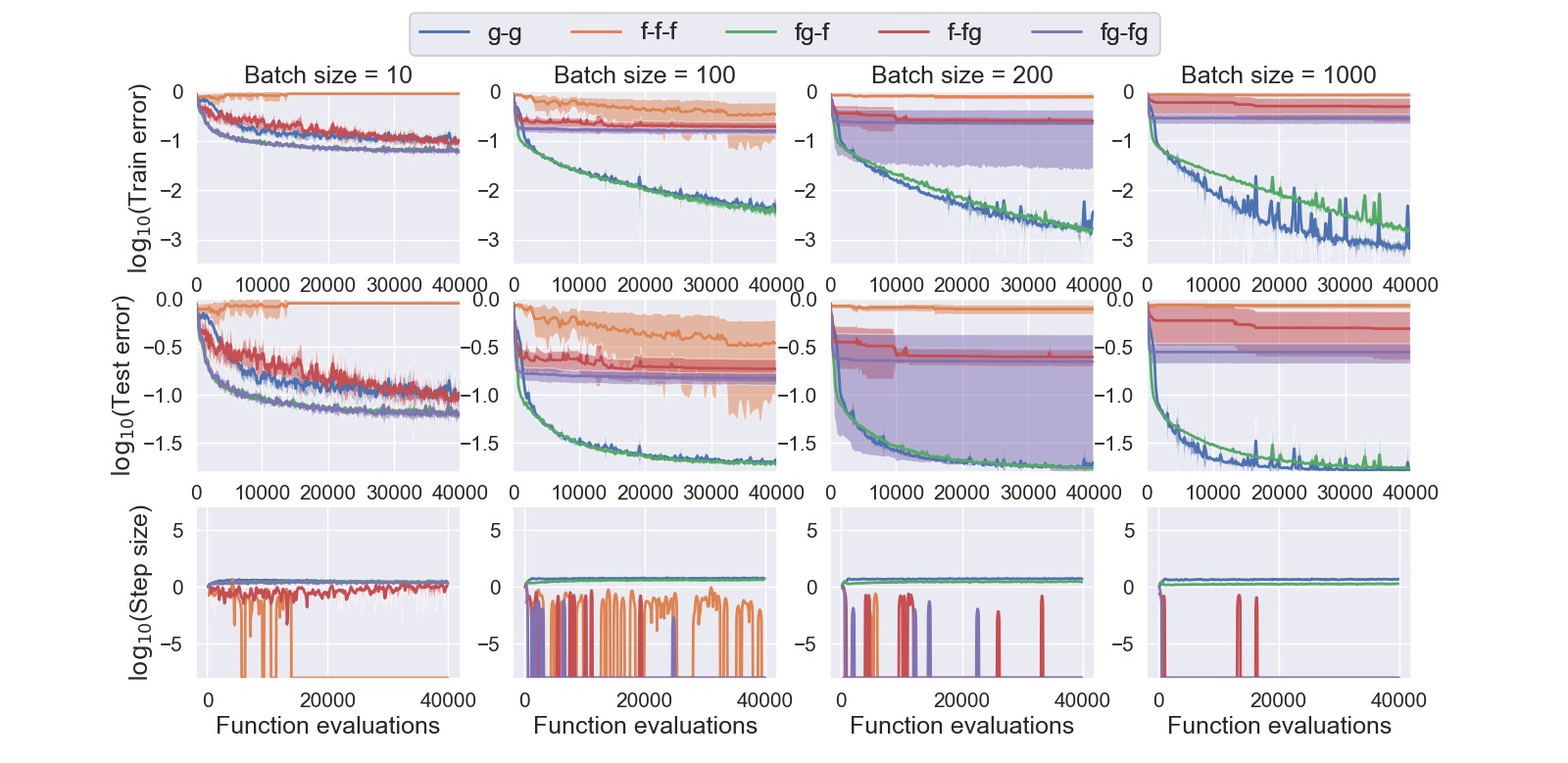}
	\caption{
		Enforced bound limits: comparison of approximation-assisted line search methods with various information; 1) derivative-only (g-g), 2) function-value-only (f-f-f), 3) mixed approximation with function values at both $ \alpha_{0} $ and $ \alpha_{1} $ and directional derivative only at $ \alpha_{0} $ (fg-f), 4) directional derivative only at $ \alpha_{1} $ (f-fg) and 5) directional derivatives at both $ \alpha_{0} $ and $ \alpha_{1} $ (fg-fg)  on MNIST dataset with N-II architecture with respect to function evaluations. 
	}\label{prob_d1n2_1}
\end{figure}

\begin{figure}
	\centering
	\includegraphics[width=1\linewidth]{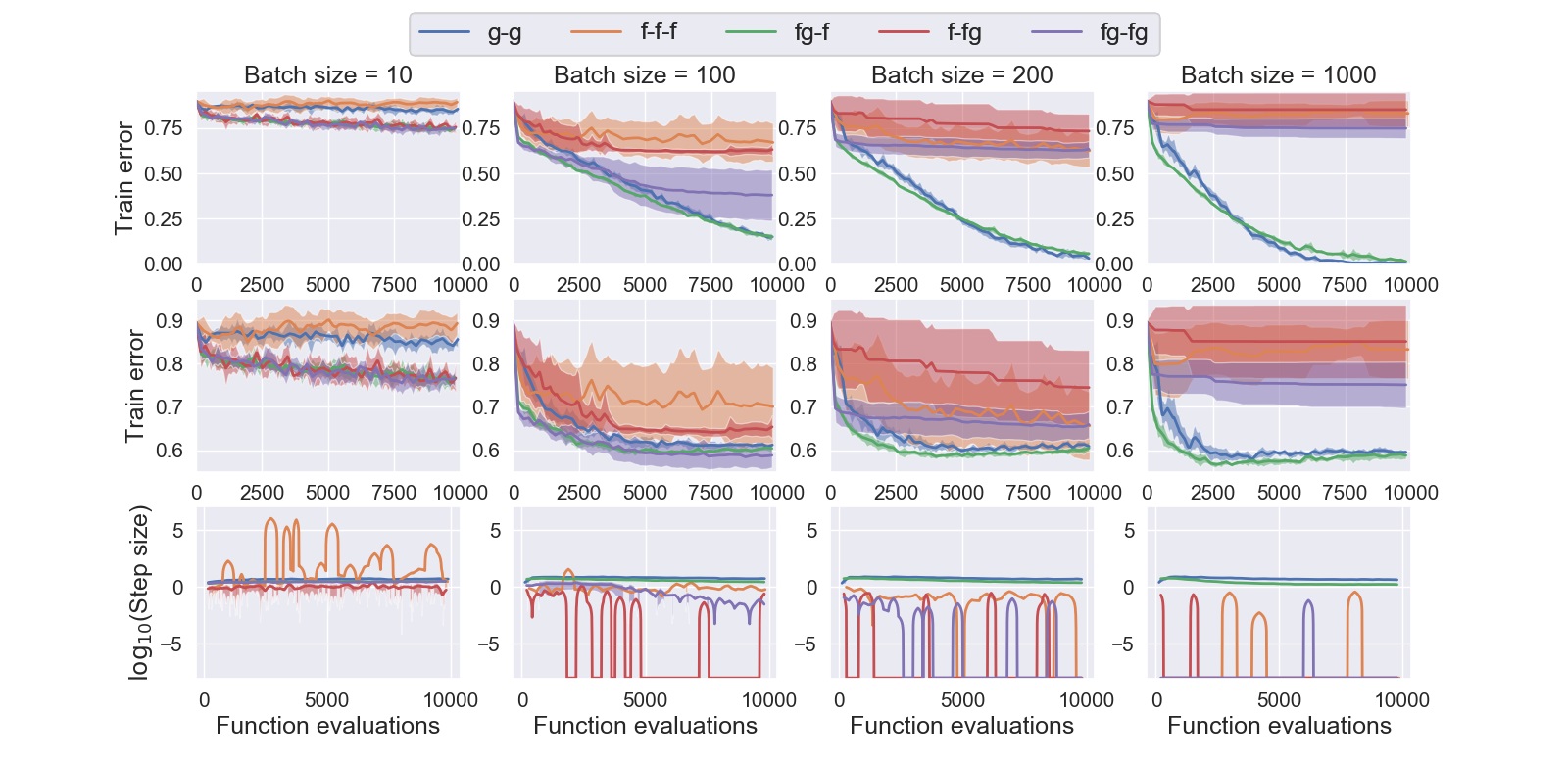}
	\caption{
		Enforced bound limits: comparison of approximation-assisted line search methods with various information; 1) derivative-only (g-g), 2) function-value-only (f-f-f), 3) mixed approximation with function values at both $ \alpha_{0} $ and $ \alpha_{1} $ and directional derivative only at $ \alpha_{0} $ (fg-f), 4) directional derivative only at $ \alpha_{1} $ (f-fg) and 5) directional derivatives at both $ \alpha_{0} $ and $ \alpha_{1} $ (fg-fg)  on CIFAR-10 dataset with N-II architecture with respect to function evaluations. 
	}\label{prob_d2n2_1}
\end{figure}

The training results with the lower and upper bounds ($ \alpha_{min} $ and $ \alpha_{max} $) activated for N-I with MNIST and CIFAR-10 are shown in Figures~\ref{prob_d1n1_1} and \ref{prob_d2n1_1}, and N-II with MNIST and CIFAR-10 are shown in Figures~\ref{prob_d1n2_1} and \ref{prob_d2n2_1}. The N-I results show that the f-f-f approximation, which do not use directional derivative information still performs unstably although the bound limits are enforced. Because we restrict the location of the initial guess $ \alpha_{1} $, any poorer performance in comparison to the no bound limit results shown in Figures~\ref{prob_d1n1_0}-\ref{prob_d1n2_0} and \ref{prob_d2n2_0} would imply how sensitive the location of information is for the approximations. This phenomenon is demonstrated both in N-I and N-II. The f-fg approximation, which used to be one of the top performing approximations in no bound limit test as illustrated in Figures~\ref{prob_d1n1_0} and \ref{prob_d2n1_0}, shows slightly underperforming results in this N-I test. Moreover, both f-fg and fg-fg approximations for the N-II results of the no bound limit tests shown in Figures~\ref{prob_d1n2_0} and \ref{prob_d2n2_0} used to be competitive to the best performing approximations, g-g and fg-f. However, the enforced bound limit tests shown in Figures~\ref{prob_d1n2_1}-\ref{prob_d2n2_1} show that the approximation error of the f-fg and fg-fg approximations are sensitive to the location of the information captured.


The investigations up to this point indicates that it is critical for the approximations to enforce the directional derivative $ f'_{0} $ at the starting point $ \alpha_{0} =0 $. This significantly decreases the variance of $ \alpha_{*} $. Moreover, the approximations with less but selective data such as the g-g and fg-f approximations are more stable and competitive than the performance of the f-f-f and fg-fg approximations, which utilize more but higher variance data. 

\subsection{Fixed batch training comparison}
\begin{figure}
	\centering
	\subfloat[MNIST N-I]{%
		\resizebox*{7cm}{!}{\includegraphics{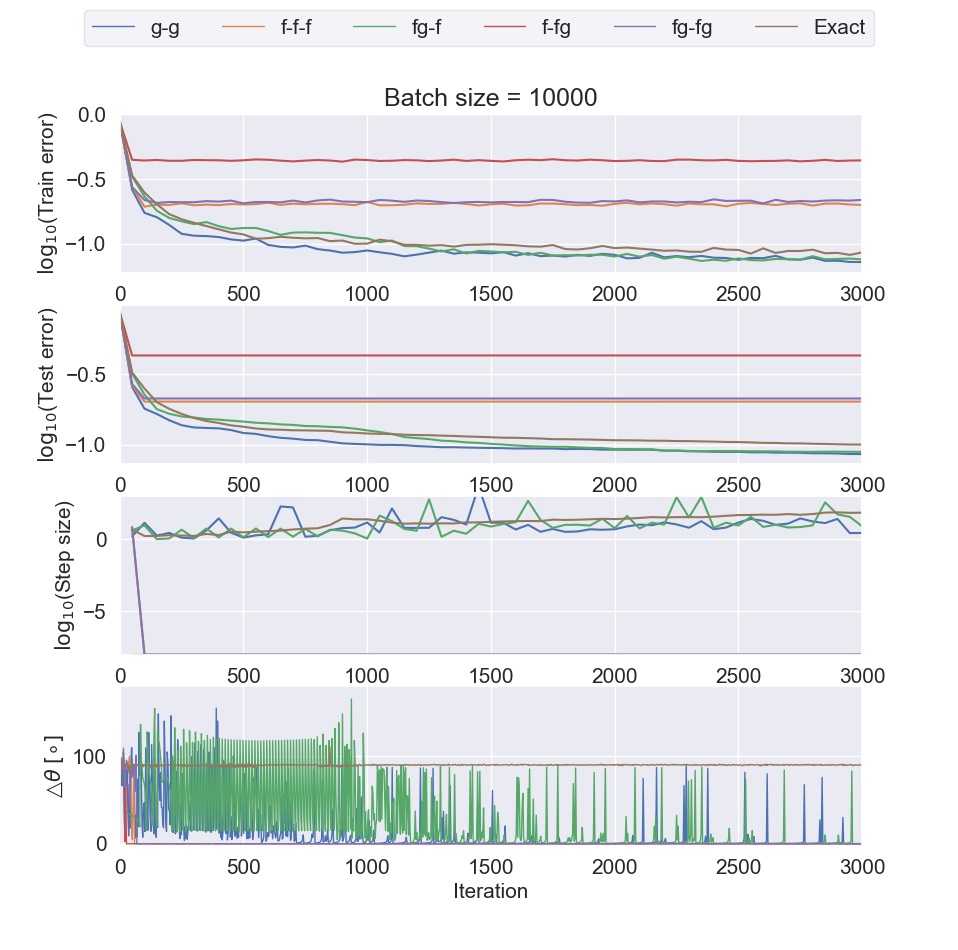}}}\hspace{5pt}
	\subfloat[CIFAR-10 N-I]{%
		\resizebox*{7cm}{!}{\includegraphics{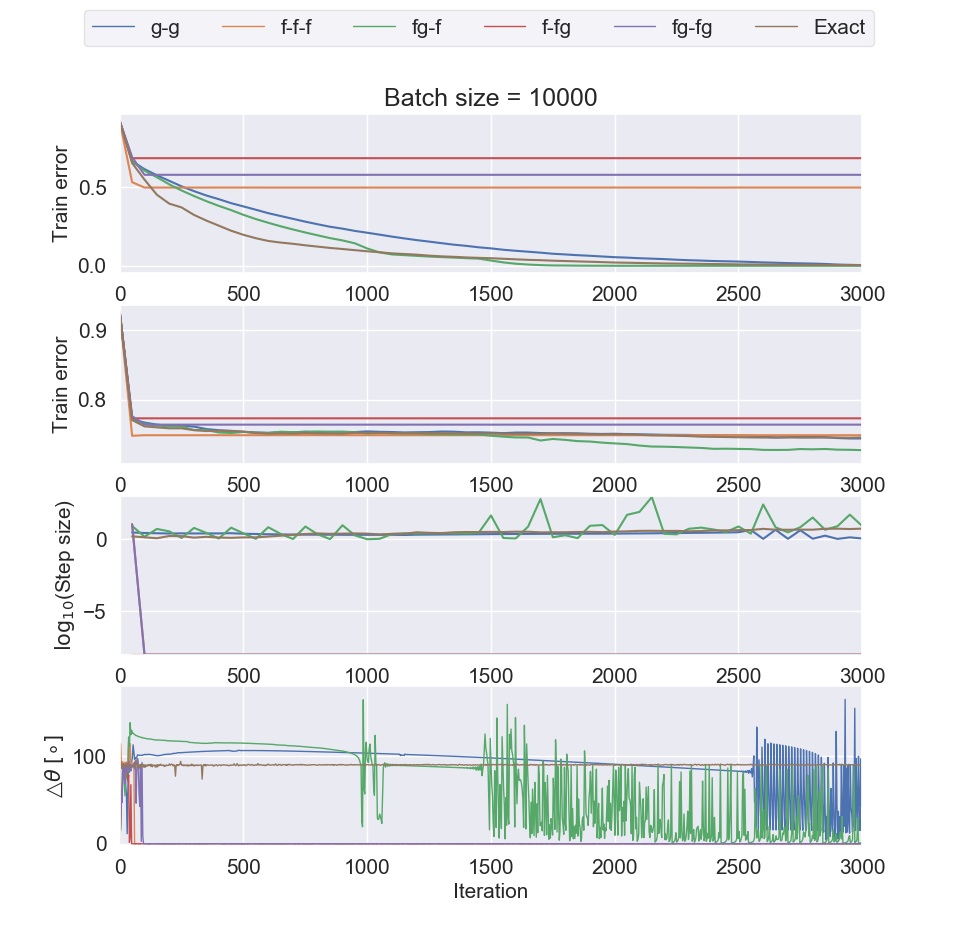}}} \hspace{5pt}
	\subfloat[MNIST N-II]{%
		\resizebox*{7cm}{!}{\includegraphics{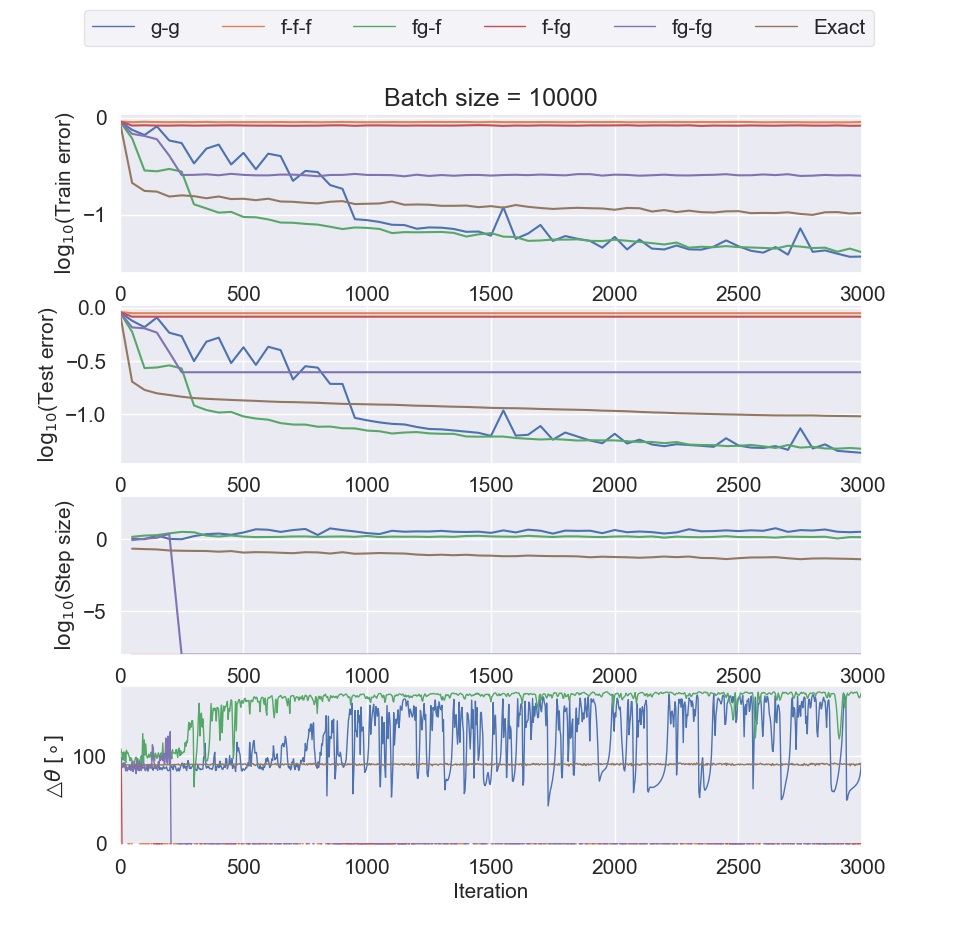}}}\hspace{5pt}
	\subfloat[CIFAR-10 N-II]{%
		\resizebox*{7cm}{!}{\includegraphics{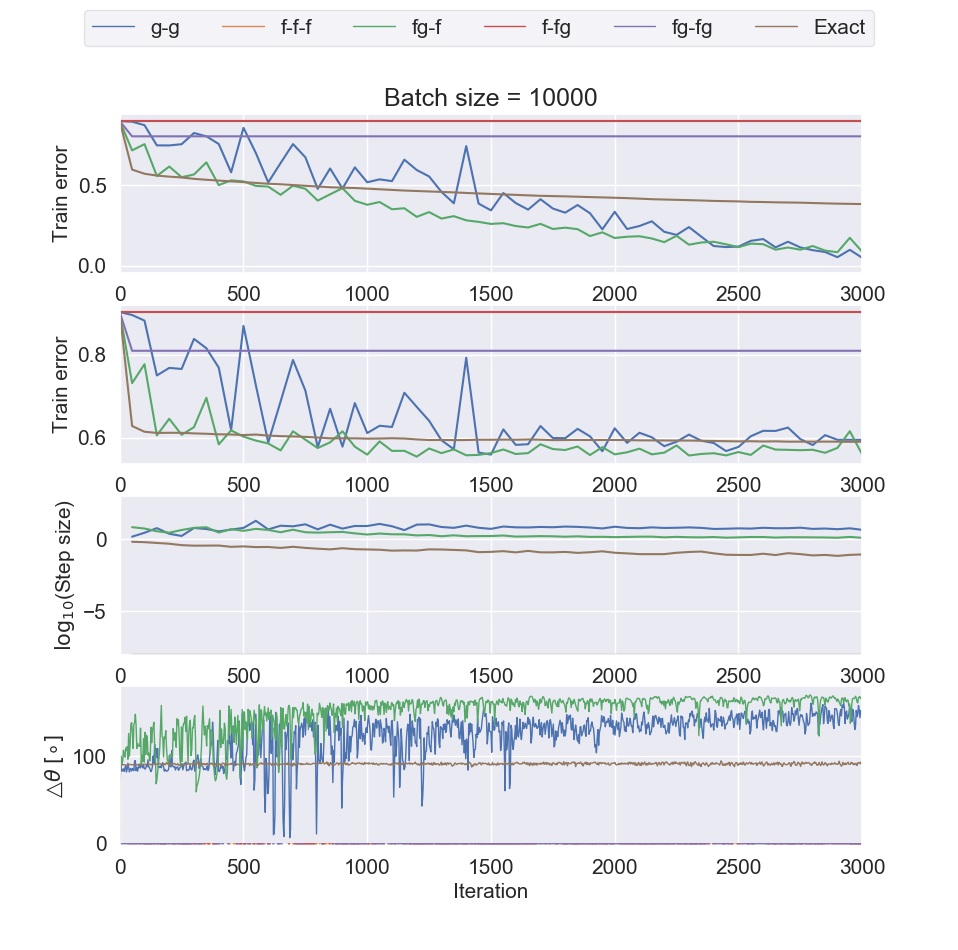}}}\hspace{5pt}
	\caption{Comparison of approximation-assisted line search methods against the exact line search; 1) derivative-only (g-g), 2) function-value-only (f-f-f), 3) mixed approximations with function values at both $ \alpha_{0} $ and $ \alpha_{1} $ and directional derivative only at $ \alpha_{0} $ (fg-f), 4) directional derivative only at $ \alpha_{1} $ (f-fg), 5) directional derivatives at both $ \alpha_{0} $  and $ \alpha_{1} $ (fg-fg) and 6) exact line search  on N-I with (a) MNIST, (b) CIFAR-10 and N-II with (c) MNIST, (d) CIFAR-10 with respect to iterations.} \label{fixed}
\end{figure}

The last experimental results shown in Figure~\ref{fixed} present the results of the line search methods using a fixed larger batch, $ m = 10000 $ for the N-I with (a) MNIST and (b) CIFAR-10 and the N-II with (c) MNIST and (d) CIFAR-10 classification tasks for the first 3000 iterations. Note that since the size of the full train samples of the CIFAR-10 is 10000, this implies the full batch training for the CIFAR-10 dataset. Since the batch remains fixed during training, the sampling error is constant, and we have a smooth loss function given our neural network architectures, that an exact line search can find at every iteration. The results of the approximations were compared to the exact golden section line search \cite{nocedal2006numerical}. We compare results with respect to the number of iterations instead of function evaluations.

Along the rows, the train error, test error, step size and the measured angles $ \bigtriangleup\theta $ between the current and previous search direction  in degrees are shown in Figure~\ref{fixed}. The N-I results of the MNIST and CIFAR-10 datasets in Figure~\ref{fixed}(a) and (b) show that whereas the f-f-f, f-fg and fg-fg approximations converge to relative high errors, the fg-f and g-g approximations show competitive results to the exact line search. The step size of the 2 approximations are also similar to that of the exact line search. Note that the difference in angle $ \bigtriangleup\theta $ between the previous and current search direction for the exact line search is always orthogonal, $ 90^{\circ} $ because the search direction is the gradient descent direction but the $ \bigtriangleup\theta $ of the 2 approximations alternate around the orthogonality. This means that the approximations create some conjugacy in the search directions, and this also means that they alternate between overshooting and undershooting due to approximation errors as shown in Figure~\ref{shooting}.

The N-II results for MNIST and CIFAR-10 shown in Figures~\ref{fixed}(c) and (d) demonstrate that using approximation-assisted line search (fg-f and g-g) helps to find the lower minima than the solution found using the exact line search although the initial convergence of them is slower. The average step sizes of the fg-f and g-g approximations are slightly larger than that of the exact line search while the $ \bigtriangleup\theta $ is larger than $ 90^{\circ}$ due to the larger step size analysis explained early in Figure~\ref{contour}. Note that $ \bigtriangleup\theta $ of the 2 approximations are continuously increasing, and this can be seen as exploitation on flatter surface close to the end of training.

\begin{figure}
	\centering
	\subfloat[f-f-f]{%
		\resizebox*{4cm}{!}{\includegraphics{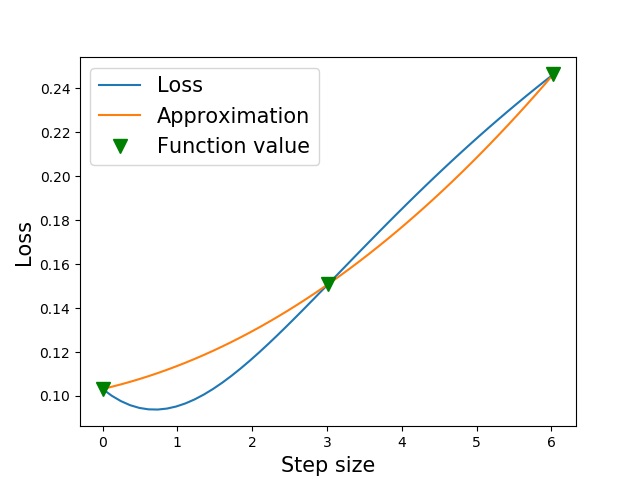}}}\hspace{5pt}
	\subfloat[f-fg]{%
		\resizebox*{4cm}{!}{\includegraphics{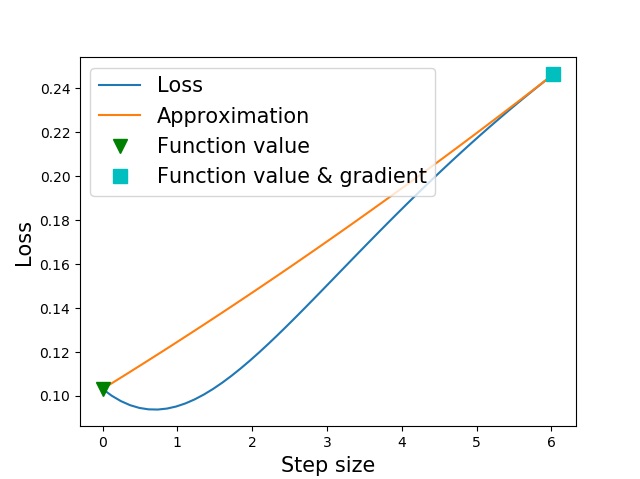}}} \hspace{5pt}
	\subfloat[fg-fg]{%
		\resizebox*{4cm}{!}{\includegraphics{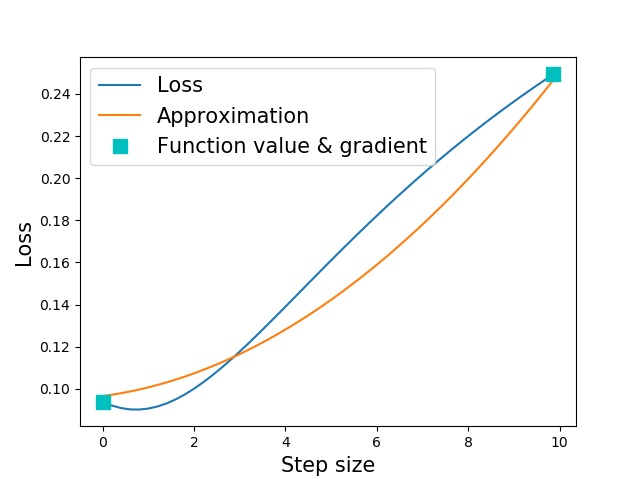}}}\hspace{5pt}
	\caption{Illustration of poor approximations constructed for fixed batch training problem: (a) f-f-f, (b) f-fg and (c) fg-fg approximations for N-II with the CIFAR-10 dataset.} \label{poor}
\end{figure}

Although this was a continuous loss function, it is noticed that the f-fg, f-f-f and fg-fg approximations fail both on the N-I and N-II tasks because of the high bias of the 2nd order polynomial approximations. Figure~\ref{poor} shows the examples of the poor f-f-f, f-fg and f-f-f approximations which provide unacceptable solutions (i.e. $ \alpha < 0 $) due to poor data provided for constructing the approximations. Hence, the step size minimum bound $ \alpha_{min} $ is chosen instead which leads to no visible improvement during training. Since the f-f-f and f-fg approximations do not have directional derivative information at $ \alpha_{0} $, they do not guarantee that $ \alpha >0 $. The fg-fg approximation, even with the directional derivative provided, may still construct the approximation which is biased towards other information, and cause failures ($ \alpha < 0 $) as shown in Figure~\ref{poor}(c). This clearly motivates that using less but more relevant data such as directional derivative information helps reducing the approximation errors for line search, as we can observe from the overall performance of the fg-f and g-g.

\section{Conclusion}
Mini-batch sub-sampling (MBSS) within the context of line searches can be either static or dynamic.  Static or dynamic sub-sampling refers to whether the mini-batch is updated per loss function evaluation (dynamic) or per search direction evaluation (static) where it is kept fixed for the loss function evaluations along that search direction. Dynamic MBSS allows for observing more samples in less number iterations when compared to static MBSS while keeping the computational memory limits. However, it computes point-wise discontinuous loss functions with stochastic loss and directional derivatives. 

As a result, it is challenging to conduct line search with dynamic MBSS due to the discontinuities. In the stochastic environment, the line search using function values finds multiple local minima. However, if we search the sign changes of directional derivatives, we can find the narrower ranges of local minima compared to searching for lower function values. Enforcing selective  information (e.g. function value and directional derivative), we constructed five 1D quadratic approximations to resolve step sizes along descent directions.

The empirical results indicate that directional derivative information is much more critical than function value information. Especially, the directional derivative information at the starting point is more informative than at other points because it guarantees that the solution is along the descent search direction, which also helps to reduce the variance of the solutions within this stochastic setting. It was found that any information for small batch sizes is useful, but for deeper network tasks with only slightly larger mini-batch sizes, the distinction between fewer samples of selective information is much more beneficial than larger numbers of indiscriminate information. This implies that having less information, that  focuses on directional derivatives outperforms approximations using more information. 

\section*{Acknowledgment}
This research was supported by the National Research Foundation (NRF), South Africa and the Center for Asset Integrity Management (C-AIM), Department of Mechanical and Aeronautical Engineering, University of Pretoria, Pretoria, South Africa, and we would like to express our special thanks to Nvidia Corporation for supplying the GPUs on which this research was conducted.

\bibliographystyle{tfs}
\bibliography{bib_info}

\newpage

\appendix
\section{Pseudo code for various approximations}
\label{appendix}

\begin{algorithm}[H]
	\DontPrintSemicolon 
	\KwIn{$ \alpha_{1} $, $ \hat{f}_{0} $, $ \hat{f}_{1} $, $ \hat{f}'_{0} $, $ \epsilon_{k} $}
	\KwOut{$\alpha_{*}$}
	$ \alpha_{*} = \alpha_{1}$ \;
	Define a matrix $ \boldsymbol{A}_{2} $ and a vector $ \boldsymbol{b}_{2} $ from (\ref{fgf})\;
	\If{$ rank(\boldsymbol{A}) = 3 $}{
		
		Solve for the constants $	\boldsymbol{k} = \boldsymbol{A}^{-1}\boldsymbol{b} $ from (\ref{quad_func})\;
		\If{$ k_{1}  > \epsilon_{k} $}{
			$\alpha_{*} =   -k_{2}/(2k_{1}) $\;
			\uIf{$ \alpha_{*} \geq \alpha_{max} $}{$\alpha_{*} = \alpha_{max}$}
			\ElseIf{$ \alpha_{*} \leq \alpha_{min} $}{$\alpha_{*} = \alpha_{min}$}
		}
	}
	\caption{\texttt{StepSizeFGF}}
	\label{stepFGF}
\end{algorithm}

\begin{algorithm}[H]
	\DontPrintSemicolon 
	\KwIn{$ \alpha_{1} $, $ \hat{f}_{0} $, $ \hat{f}_{1} $, $ \hat{f}'_{1} $, $ \epsilon_{k} $}
	\KwOut{$\alpha_{*}$}
	$ \alpha_{*} = \alpha_{1}$ \;
	Define a matrix $ \boldsymbol{A}_{3} $ and a vector $ \boldsymbol{b}_{3} $ from (\ref{ffg})\;
	\If{$ rank(\boldsymbol{A}) = 3 $}{
		
		Solve for the constants $	\boldsymbol{k} = \boldsymbol{A}^{-1}\boldsymbol{b} $ from (\ref{quad_func})\;
		\If{$ k_{1}  > \epsilon_{k} $}{
			$\alpha_{*} = -k_{2}/(2k_{1}) $\;
			\uIf{$ \alpha_{*} \geq \alpha_{max} $}{$\alpha_{*} = \alpha_{max}$}
			\ElseIf{$ \alpha_{*} \leq \alpha_{min} $}{$\alpha_{*} = \alpha_{min}$}
		}
	}
	\caption{\texttt{StepSizeFFG}}
	\label{stepFFG}
\end{algorithm}

\begin{algorithm}[H]
	\DontPrintSemicolon 
	\KwIn{$ \alpha_{1} $, $ \hat{f}_{0} $, $ \hat{f}_{1} $, $ \hat{f}'_{0} $, $ \hat{f}'_{1} $, $ \epsilon_{k} $}
	\KwOut{$\alpha_{*}$}
	$ \alpha_{*} = \alpha_{1}$ \;
	Define a matrix $ \boldsymbol{A}_{4} $ and a vector $ \boldsymbol{b}_{4} $ from (\ref{fgfg})\;
	\If{$ rank(\boldsymbol{A}) = 3 $}{
		
		Solve for the constants $	\boldsymbol{k} = \boldsymbol{A}^{-1}\boldsymbol{b} $ from (\ref{quad_func})\;
		\If{$ k_{1}  > \epsilon_{k} $}{
			$\alpha_{*} = -k_{2}/(2k_{1}) $\;
			\uIf{$ \alpha_{*} \geq \alpha_{max} $}{$\alpha_{*} = \alpha_{max}$}
			\ElseIf{$ \alpha_{*} \leq \alpha_{min} $}{$\alpha_{*} = \alpha_{min}$}
		}
	}
	\caption{\texttt{StepSizeFGFG}}
	\label{stepFGFG}
\end{algorithm}

\newpage
\begin{algorithm}[H]
	\DontPrintSemicolon 
	\KwIn{$ \alpha_{1} $, $ \hat{f}'_{0} $, $ \hat{f}'_{1} $, $ \epsilon_{k} $}
	\KwOut{$\alpha_{*}$}
	$ \alpha_{*} = \alpha_{1}$ \;
	Define a matrix $ \boldsymbol{A}_{5} $ and a vector $ \boldsymbol{b}_{5} $ from (\ref{gg})\;
	\If{$ rank(\boldsymbol{A}) = 2 $}{
		
		Solve for the constants $	\boldsymbol{k} = \boldsymbol{A}^{-1}\boldsymbol{b} $ from (\ref{lin_func})\;
		\If{$ k_{1}  > \epsilon_{k} $}{
			$\alpha_{*} = -k_{2}/(2k_{1}) $\;
			\uIf{$ \alpha_{*} \geq \alpha_{max} $}{$\alpha_{*} = \alpha_{max}$}
			\ElseIf{$ \alpha_{*} \leq \alpha_{min} $}{$\alpha_{*} = \alpha_{min}$}
		}
	}
	\caption{\texttt{StepSizeGG}}
	\label{stepGG}
\end{algorithm}

\end{document}